\documentclass[journal]{IEEEtran}
\usepackage{algorithm}
\usepackage{algpseudocode}
\usepackage{times}
\usepackage{epsfig}
\usepackage{graphicx}
\usepackage{amsmath}
\usepackage{amssymb}
\usepackage{bm}
\usepackage{multirow}
\usepackage[utf8]{inputenc} 
\usepackage[T1]{fontenc}    
\usepackage{url}            
\usepackage{booktabs}       
\usepackage{amsfonts}       
\usepackage{nicefrac}       
\usepackage{microtype}      
\usepackage{color, colortbl}
\usepackage{color}
\usepackage{cite}
\usepackage{graphicx}
\usepackage{amsmath,amssymb}
\usepackage{acronym}
\usepackage{times}
\usepackage{multirow}
\usepackage{hyperref}
\usepackage{makecell}
\usepackage{caption}

\definecolor{Gray}{gray}{0.9}
\newcolumntype{g}{>{\columncolor{Gray}}c} 

\ifCLASSINFOpdf

\else

\fi

\begin{document}

\title{Recent Advances in Monocular 2D and 3D Human \\Pose Estimation:
A Deep Learning Perspective}

\author{Wu~Liu,~\IEEEmembership{Member,~IEEE}, Qian~Bao, Yu~Sun,
        and~Tao~Mei,~\IEEEmembership{Fellow,~IEEE}
\thanks{All authors are with JD AI Research, JD.com, Beijing, China, 100101. Email: {\{liuwu1, baoqian, tmei\}}@jd.com, yusun@stu.hit.edu.cn}.}

%
%
\markboth{Journal of \LaTeX\ Class Files,,~Vol.~x, No.~x, ~2021}%
{Shell \MakeLowercase{\textit{et al.}}: IEEE Transactions on Multimedia}
%



\maketitle

\begin{abstract}
Estimation of the human pose from a monocular camera has been an emerging research topic in the computer vision community with many applications. Recently, benefited from the deep learning technologies, a significant amount of research efforts have greatly advanced the monocular human pose estimation both in 2D and 3D areas. Although there have been some works to summarize the different approaches, it still remains challenging for researchers to have an in-depth view of how these approaches work. In this paper, we provide a comprehensive and holistic 2D-to-3D perspective to tackle this problem. We categorize the mainstream and milestone approaches since the year 2014 under unified frameworks. By systematically summarizing the differences and connections between these approaches, we further analyze the solutions for challenging cases, such as the lack of data, the inherent ambiguity between 2D and 3D, and the complex multi-person scenarios. We also summarize the pose representation styles, benchmarks, evaluation metrics, and the quantitative performance of popular approaches. Finally, we discuss the challenges and give deep thinking of promising directions for future research. We believe this survey will provide the readers with a deep and insightful understanding of monocular human pose estimation.

\end{abstract}

\begin{IEEEkeywords}
Survey for human pose estimation, deep learning, 2D and 3D pose, monocular images.
\end{IEEEkeywords}

%
\IEEEpeerreviewmaketitle

\section{Introduction}
\subsection{Motivation}

\IEEEPARstart{M}{onocular} human pose estimation (MHPE) is a fundamental and challenging task in the computer vision community. It aims to predict the human pose information, such as the spatial locations of body joints and/or the body shape parameters, from a monocular image or video. MHPE has been widely exploited for many computer vision tasks, such as person re-identification~\cite{8237689,8099586}, human parsing~\cite{nie2018mutual,Zhang_2020_CVPR}, human action recognition~\cite{7410725,8237664}, and human-computer interaction~\cite{Fang_2018_ECCV,Li_2019_CVPR}, etc. As MHPE does not need the complex multi-cameras or wearable marker points, it has become a significant part of many real-world applications, such as virtual reality, 3D movie making/editing, self-driving, motion and activity analysis, and human-robot interaction.  

    \begin{figure}[!t]
	\centering
	\includegraphics[width=0.45\textwidth]{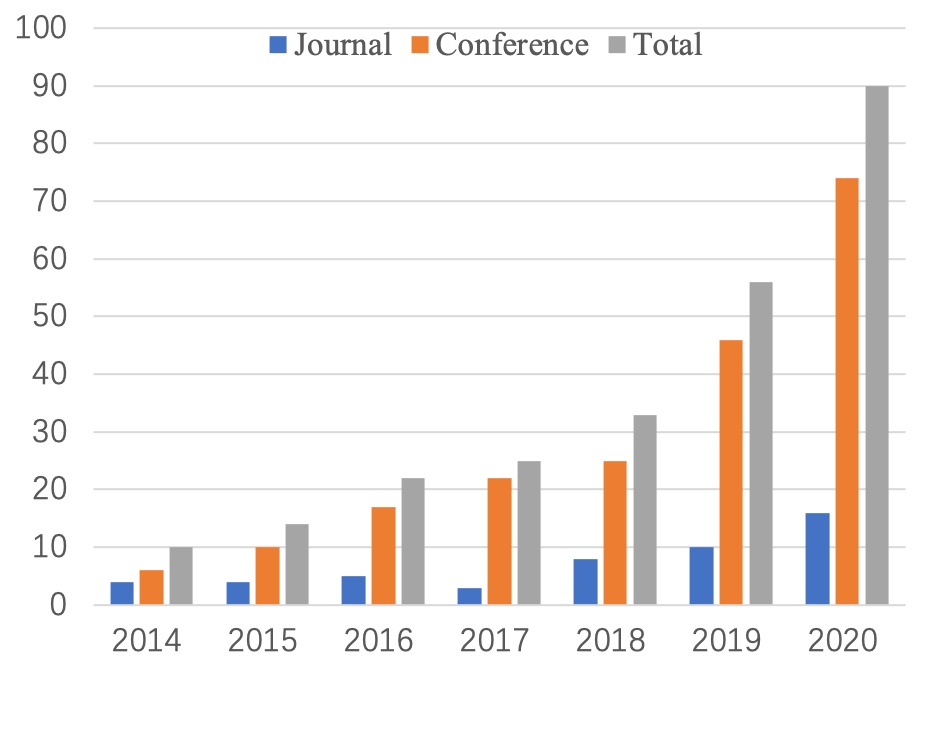}
	\caption{The number of the published papers in mainstream computer vision, multi-media, and  computer graphics conferences (CVPR, ICCV, ECCV, etc) and journals  (TPAMI, TIP, TOG, etc) from the year 2014 to 2020.}
	\label{fig:paper_number}
    \end{figure}

According to the spatial dimension of the output results, the mainstream MHPE tasks can be divided into two categories, 2D pose estimation, and 3D pose estimation. Monocular 2D human pose estimation, also known as 2D keypoint detection, aims to locate the 2D coordinates of human anatomical keypoints (body joints) from images. Considering the number of people in a given image, the 2D human pose estimation task can be further classified into single person and multi-person pose estimation. Furthermore, given a video sequence, 2D pose estimation can exploit temporal information to boost keypoint prediction in a video system. Different from solely predicting 2D locations of body joints, 3D pose estimation further predicts the depth information for more accurate spatial representation. In this process, 2D pose estimation can be exploited as the intermediate representation for 3D pose estimation. In recent years, demanding for understanding the detailed pose information of humans has driven 3D pose estimation towards predicting not only the 3D location but also the detailed 3D shape and body texture. 

Limited by data and computational resources, early research mainly focused on designing handcrafted features or fitting the deformable human body models with optimization algorithms. Recently, with the increase of large-scale 2D/3D pose datasets (e.g. COCO~\cite{Dataset_COCO}, MPII~\cite{Dataset_MPII}, Human3.6M~\cite{h36m}, and 3DPW~\cite{3dpw}), deep learning technologies have significantly boosted the performance of human pose estimation both in accuracy and efficiency. As shown in Fig.~\ref{fig:paper_number}, from the year 2014 to 2020, the number of 
published papers in the mainstream conferences (CVPR, ICCV, ECCV, etc) and journals (TPAMI, TIP, TOG, etc) in the area of computer vision, multi-media, and computer graphics has rapidly increased. 
Recent works mainly focus on network design and optimization~\cite{14SP_ToshevS14,16SP_NewellYD16,16BU_pishchulin2016deepcut, 18MP_xiao2018simple, 18MP_chen2018cascaded, 19MP_HrnetXLW19,pavlakos2017coarse,hmr,sun2018integral,kolotouros2019spin}, multitask interaction~\cite{17BU_he2017mask, 18SP_LuvizonPT18,Guler_2019_CVPR,18BU_papandreou2018personlab,18BU_kocabas2018multiposenet}, body model exploration~\cite{zhao2019semantic, smpl, puppet_zuffi2015stitched,totalcapture}, etc.

    \begin{figure*}[!t]
	\centering
	\includegraphics[width=1.0\textwidth]{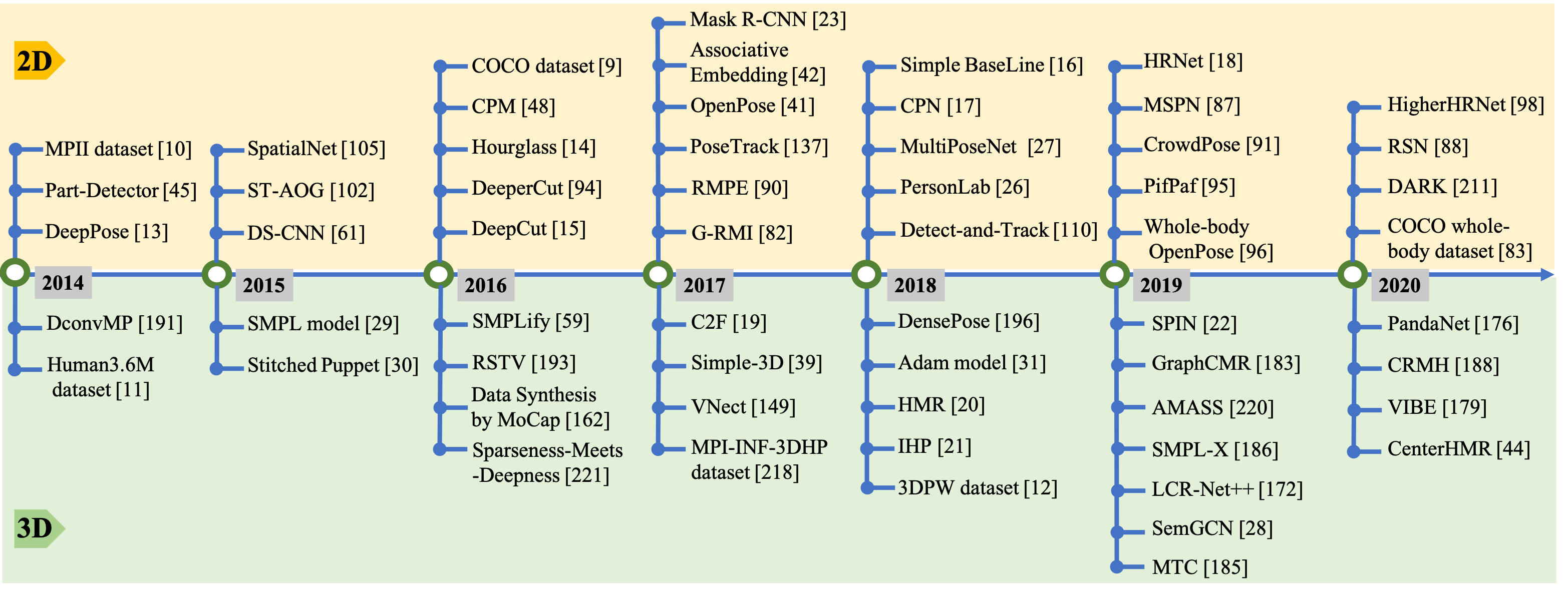}
	\caption{Milestones, idea or dataset breakthroughs, and the state-of-the-art methods for 2D (top) and 3D (bottom) pose estimation from the year 2014 to 2021.}
	\label{fig:timeline}
    \end{figure*}

Although great successes have been achieved in performance and practice, few works have comprehensively reviewed the representative algorithms or given insightful analyses of 2D-to-3D pose estimation. 
On one hand, some previous surveys~\cite{12_3Dsurvey, 15_2Dsurvey,16_2Dsurvey,2dposereview1} reviewed traditional methods, such as body models or handcrafted features, without recent deep-learning-based approaches. 
On the other hand, recent surveys have mainly focused on one aspect of either 2D pose estimation~\cite{20_2Dsurvey} or 3D pose estimation~\cite{16_3Dsurvey}, without a comprehensive perspective to explore the intrinsic connections between 2D and 3D.
The survey~\cite{survey2020monocular} describes recent works of representative 2D pose estimation methods and a few 3D pose estimation methods up to the year 2019. However, it does not well summarize the relative 3D pose and shape estimation methods, and neglects the perspective from 2D to 3D. Therefore, a more comprehensive survey covering the recent advantage of pose estimation is of great need in this community. 

In this paper, we provide a comprehensive review of the deep learning-based MHPE approaches from 2D to 3D in recent years. We believe that most representative MHPE methods have intrinsic similarities and connections.  
Moreover, with the rapid development of 3D pose and shape estimation, it is necessary to have a deeper survey on the human pose estimation from 2D to 3D. 
Therefore, compared with the paper~\cite{survey2020monocular}, our survey has the following differences and advantages. 1) We summarize the prevailing networks for both 2D and 3D pose estimation in the unified frameworks. They represent the representative paradigms. 2) We provide insightful analyses for human 3D representation, 3D datasets, 3D shape recovery methods, as well as the challenges and further work for 3D pose estimation. 3) Besides, we released a detailed code toolbox\footnote{https://github.com/Arthur151/SOTA-on-monocular-3D-pose-and-shape-estimation} for 3D pose data processing, which will be timely and useful for 3D pose research.
We summarize a timeline in Fig.~\ref{fig:timeline}, which shows the milestones, idea or dataset breakthroughs, and the state-of-the-art methods for 2D and 3D pose estimation from the year 2014 to 2021. We can see that new approaches and new datasets promote each other. 2D pose estimation has achieved explosive development since 2016 with breakthroughs both in ideas and datasets. Meanwhile, 3D pose estimation has also developed rapidly in recent years.

\subsection{Overview of Deep Learning Framework for MHPE}

The human body is nonrigid and flexible for high degree-of-freedom poses, therefore, predicting human pose estimation from a monocular camera faces many challenges, such as complex or strange posture, person-object/person-person interaction or occlusion, and crowded scenes, etc. Different camera views and complex scenes will also introduce problems of truncation, image blur, low resolution, and small target persons.

    \begin{figure*}[!t]
	\centering
	\includegraphics[width=0.85\textwidth]{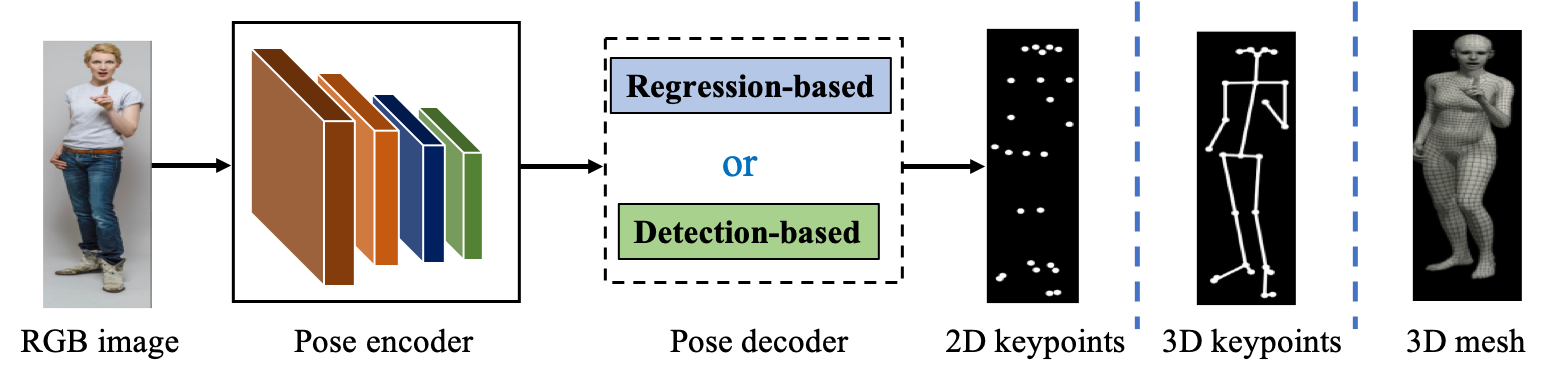}
	\caption{Typical framework for single person pose estimation.}
	\label{fig:outline}
    \end{figure*}

    \begin{figure*}[!t]
	\centering
	\includegraphics[width=0.8\textwidth]{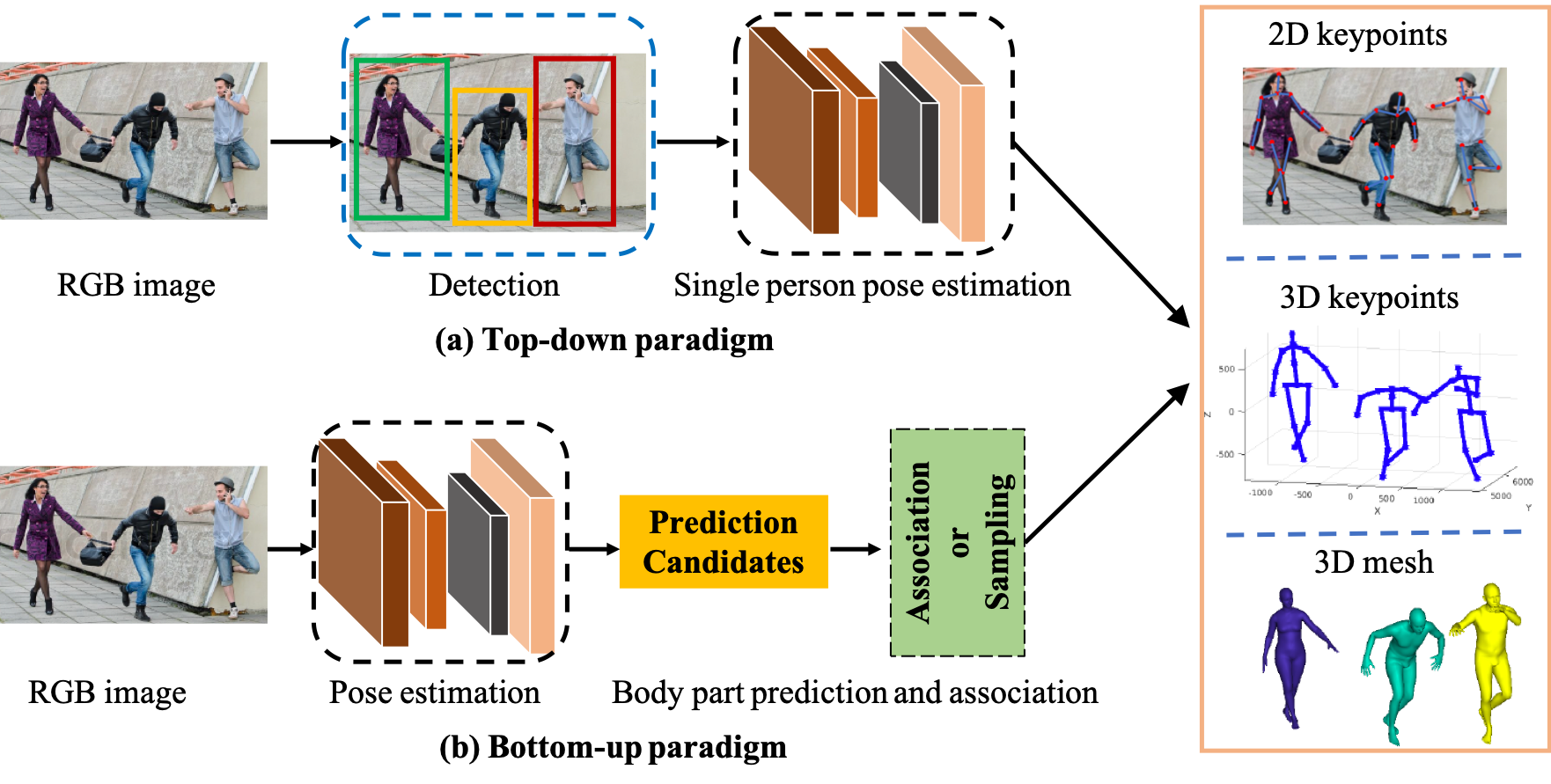}
	\caption{Typical frameworks for multi-person pose estimation.}
	\label{fig:outline_multiperson}
    \end{figure*}

To address these problems, existing methods explore the powerful representation of deep learning to mine more clues for pose estimation. Although they are different in either global design or detailed optimization, the network architectures of milestone methods have internal similarities. As shown in Fig.~\ref{fig:outline}, most of the prevailing single person pose estimation networks~\cite{16SP_NewellYD16, 18MP_xiao2018simple,18MP_chen2018cascaded, 19MP_HrnetXLW19, martinez2017simple, sun2019dsd-satn, hmr} can be regarded as consisting of a pose encoder (also called feature extractor) followed by a pose decoder. The former aims at extracting high-level features through a high-to-low resolution process. The latter estimates the target output, 2D/3D keypoint location or 3D mesh, in a detection-based manner or a regression-based manner. For the pose decoder, detection-based methods yield feature maps or heatmaps, while regression-based methods directly output the target parameters. 
Following the unified frameworks, we describe details of network design for 2D  and 3D pose estimation in Sections III and IV, respectively. 

For multi-person scenes, to estimate the 2D or 3D pose of each person, existing works exploit the top-down paradigm or bottom-up paradigm.   
The top-down framework first detects the person areas and then extracts the bounding box-level features from them. The features are used to estimate the pose results for every single person. In contrast, the bottom-up paradigm first detects all target outputs and then assign them to different people by grouping~\cite{openpose,17BU_newell2017associative,zanfir2018deep} or sampling~\cite{CenterHMR}. As shown in Fig.~\ref{fig:outline_multiperson}, the representative multi-person methods of the two paradigms also rely on pose-encoder-and-decoder-based architecture with network input being either the detected bounding box or the whole image.

Therefore, how to design an effective pose encoder and pose decoder architecture is a common and popular topic in pose estimation. Different from classification, detection, and semantic segmentation, human pose estimation needs to deal with the subtle differences between body parts, especially in the unavoidable truncation, crowded, and occluded cases.
To achieve this, the body structural models~\cite{14SP_TompsonJLB14,16SP_ChuOLW16,18SP_TangYW18}, multi-scale feature fusion~\cite{16SP_NewellYD16,19MP_HrnetXLW19}, multistage pipelines~\cite{SP16-CPM,openpose}, refinement in a coarse-to-fine manner~\cite{pavlakos2017coarse,19_ContextualRefine}, multi-task learning~\cite{17BU_he2017mask,18BU_kocabas2018multiposenet,18BU_papandreou2018personlab}, etc, have been explored and designed. We will introduce them in detail in Section III and IV. 

Moreover, regarding estimating 3D poses from monocular images, another challenge is the insufficient in-the-wild 3D training data. Because of the equipment constraints, common 3D pose datasets are often captured in restrained experimental environments. For example, the most widely used 3D pose dataset Human3.6M~\cite{h36m} contains only 15 indoor activities performed by seven persons. Therefore, the diversities of human poses, shapes, and scenes are extremely limited. Models solely trained on these datasets are prone to fail on the in-the-wild images. To address this problem, many methods take the 2D pose as the intermediate representation or extra supervision, and learn from in-the-wild 2D pose information. Nevertheless, there are inherent ambiguities in this process, i.e., a single 2D pose may correspond to multiple 3D poses and vice versa. To solve the inherent ambiguities, we must consider how to fully exploit the common structure prior to the human body, motion continuity, and multi-view consistency.


In conclusion, considering the main challenges of the task and the unified frameworks of the representative paradigms, in this paper, we systematically analyze the deep learning-based 2D and 3D MHPE approaches proposed since the year 2014. 
The rest of the paper is organized as follows. We first introduce the pose estimation background in Section II, which is fundamental for understanding the MHPE task. Then in Section III, we introduce the representative approaches for 2D pose estimation, including single person pose estimation, multi-person pose estimation, pose estimation in videos, and the related tasks of 2D pose estimation, respectively. Then in Section IV, for 3D pose estimation, we detail the approaches according to their motivations and challenges. In addition, we introduce widely used 2D and 3D pose benchmarks in Section V and compare their state-of-the-art methods. Finally, in Section VI, we conclude the paper and give some insight into future research.

	\section{Background}
	\subsection{Representations for Human Body~\label{sec:representation}}
	
    \begin{figure*}[!t]
	\centering
	\includegraphics[width=1.0\textwidth]{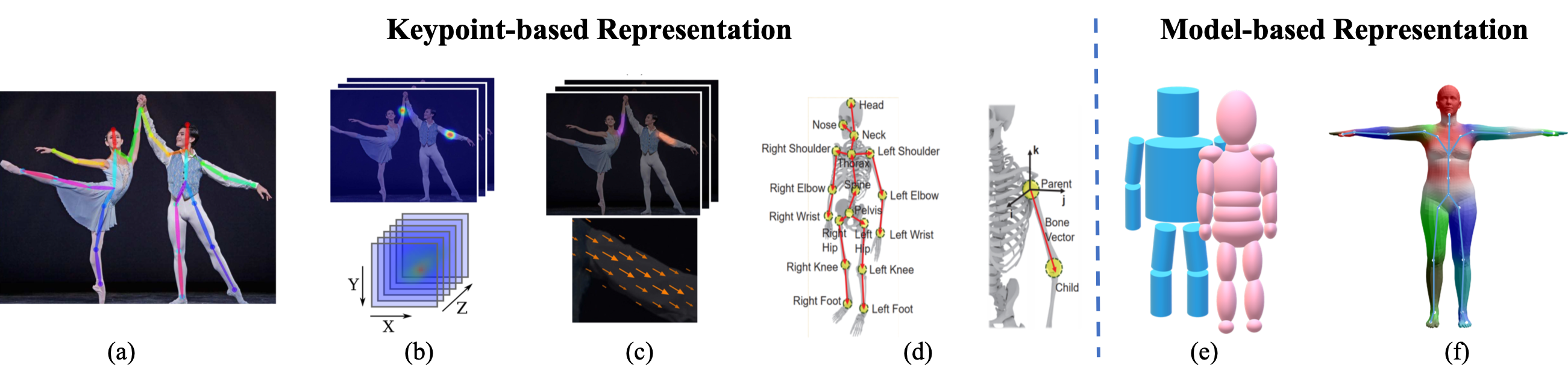}
	\caption{Widely used human body representations: (a)  2D keypoints~\cite{openpose}; (b) 2D heatmap (upper)~\cite{openpose} and volumetric heatmap (below)~\cite{18SP_LuvizonPT18}); (c) orientation map PAF~\cite{openpose}; (d) hierarchical bone representation~\cite{ETD_2020_CVPR}; (e) cylinder model (blue) and ellipBody (pink); and (f) skeleton-driven skinned multi-person linear model (SMPL)~\cite{smpl}.}
	\label{fig:representation}
    \end{figure*}
    
    Various representations of the human body have been developed to describe the complex human body pose in different aspects. They have shown various characteristics to handle different challenges of pose estimation. Existing representations can be divided into two categories: 1) keypoint-based representation; and 2) model-based representation.	
\subsubsection{Keypoint-based Representation~\label{sec:keypoint representation}}
 
2D or 3D coordinates of body keypoints are the simple and intuitive representations for the body skeleton, which have several representation forms.
  
	\textbf{2D/3D keypoint coordinates.}  Body keypoints can be explicitly described by the 2D/3D coordinates. As shown in Fig.~\ref{fig:representation} (a), the keypoints are connected following the inherent body structure. The orientations of the body part can be derived from these connected limbs. 
    
    \textbf{2D/3D heatmaps.} To make the coordinates more suitable for being regressed by a convolutional neural network, many methods represent the keypoint coordinates in a heatmap manner. As shown in Fig.~\ref{fig:representation} (b), the Gaussian heatmap of each keypoint has a high response value on the corresponding 2D/3D coordinates and a low response value at other positions. 
    
    \textbf{Orientation maps.}  Some methods~\cite{openpose,luo2018orinet} take body keypoints' orientation map as the auxiliary representation of heatmaps. OpenPose~\cite{openpose} develops the well-known part affinity fields (PAFs) to represent the 2D orientation between limbs.  As shown in Fig.~\ref{fig:representation} (c), a PAF is a 2D vector field that associates two keypoints of a limb. Each pixel in the field contains a 2D vector that points from one part of the limbs to the other. Orinet~\cite{luo2018orinet} further develops it into the 3D orientation map, which can explicitly model the limb orientations.
    
    \textbf{Hierarchical bone vectors.} The 2D version of hierarchical bone representation was proposed in the compositional human pose (CHP)~\cite{sun2017compositional}, which is the combination of joints and bone vectors. Xu et al.~\cite{DKA_Xu_2020_CVPR} and Li et al.~\cite{ETD_2020_CVPR} further developed it to 3D.  As shown in Fig.~\ref{fig:representation} (d), the 3D human skeleton is represented by a set of bone vectors. Each bone vector is pointing from the parent keypoint to the child keypoint, following a kinematic tree. Each parent keypoint is associated with a local spherical coordinate system. The bone vector can be represented by a spherical coordinates in this system. 
    
\subsubsection{Model-based Representation\label{sec:smpl}}

Model-based representation is developed according to the inherent structural characteristics of the human body. It provides richer body information than the keypoint-based description. The model-based representation can be divided into the part-based volumetric model and the statistical 3D human body model.

\textbf{Part-based volumetric model.} 
Part-based volumetric models are developed to address challenges in reality. For example, in~\cite{Cheng_2019_ICCV}, the cylinder model was developed to generate the labels of occluded parts. As shown in the blue model of Fig.~\ref{fig:representation} (e), each limb is represented as a cylinder. Each cylinder is located by aligning the top and bottom surface centers with the 3D keypoints of the limb. Similarly, as shown in the pink model of Fig.~\ref{fig:representation} (e), an EllipBody model is proposed to take the ellipsoid as the basic unit of body parts~\cite{wang2020ellipbody}. It is more flexible than a cylinder. 

\textbf{Detailed statistical 3D human body model.} Compared with the part-based volumetric model, the statistical 3D human body mesh describes more detailed information including the body pose and shape. We introduce the most widely used skinned multi-person linear model (SMPL)~\cite{smpl}, which is a skeleton-driven human body model. SMPL disentangles the shape and pose of a human body, and encodes the 3D mesh into low-dimensional parameters. It establishes an efficient mapping $M( \boldsymbol{\beta}, \boldsymbol{\theta};\Phi):\mathbb{R}^{| \boldsymbol{\theta}|\times| \boldsymbol{\beta}|}\mapsto\mathbb{R}^{3 \times 6890}$ from shape $  \boldsymbol{\beta}$ and pose $ \boldsymbol{\theta}$ to a triangulated mesh with 6,890 vertices, where $\Phi$ represents the statistical prior of the human body. The shape parameter $ \boldsymbol{\beta} \in \mathbb{R}^{10}$ is the linear combination weight of 10 basic shapes. The pose parameter $ \boldsymbol{\theta} \in \mathbb{R}^{3 \times 23}$ represents the relative 3D rotation of 23 joints in the axis-angle representation. 
Then a linear regressor $ \boldsymbol{R} \in \mathbb{R}^{6890 \times 24} $ is developed to derive preselected body joints $\boldsymbol{J} \in \mathbb{R}^{3 \times 24} $ from 6890 vertices of human body mesh via $J=M( \boldsymbol{\beta}, \boldsymbol{\theta};\Phi)\boldsymbol{R} $. The linear combination operation of this regressor guarantees that joint location is differentiable with respect to shape $\boldsymbol{\beta}$ and pose $\boldsymbol{\theta} $ parameters. 
	
	\subsection{3D-to-2D Projection}
	3D-to-2D projection connects the 3D space to the 2D image plane. It is important to introduce this tool to better understand the methods that use it. 3D-to-2D projection uses a camera model to generate 3D-2D pose pairs~\cite{fang2018learning,ETD_2020_CVPR},  supervise 3D poses using 2D pose annotations~\cite{hmr,sun2019dsd-satn,keep}, or refine 2D poses via 3D pose projection~\cite{DKA_Xu_2020_CVPR}. The perspective camera model and weak-perspective camera model are two kinds of widely used camera models.
	
	\textbf{Perspective camera model.} The perspective camera model is usually used to project the points in the 3D space into 2D pixel coordinates on the image plane. Generally, it consists of two steps. First, we need to transform the 3D points into the camera coordinates using the extrinsic matrix $[\boldsymbol{R}|\boldsymbol{t}]$, which describes the camera rotation and translation. Second, we need the intrinsic matrix $\boldsymbol{K}$ to make an adaptive adjustment for accurate projection. Therefore, the 2D projection $J_{2d}$ of 3D keypoints $J_{3d}$ can be described as  $J_{2d} = \boldsymbol{K} [\boldsymbol{R}|\boldsymbol{t}] J_{3d}$.
    
    
	\textbf{Weak-perspective camera model.} In most situations, the input 2D images are un-calibrated and complete perspective camera parameters can hardly be retrieved. Therefore, the weak-perspective camera model is more widely used in most existing methods for calculating the 2D projection $J_{wp2d}$ of 3D keypoints $J_{3d}$ by $J_{wp2D} = s {\rm{\Pi}}(\boldsymbol{R}{J_{3d}})+ \boldsymbol{t}$,  where $\boldsymbol{R} \in \mathbb{R}^{3}$ is the global rotation parameter, $\rm{\Pi}$ is an orthographic projection operation, $ \boldsymbol{t} \in \mathbb{R}^{2}$ and $ s\in \mathbb{R}$ represent the translation and scale on the image plane, respectively.
    

\begin{table*}[]
\setlength\tabcolsep{4pt}
  \small{
\caption{Representative deep learning-based methods for monocular 2D pose estimation.}\label{table_all_methods}

\begin{tabular}{c|c|c|p{12cm}}
\hline
       \textbf{Image/}                       &        \textbf{Single/}                  &             \multirow{2}{*}{ \textbf{Main idea} }                                                       &      \multirow{2}{*}{\textbf{Methods}}\\ 
       \textbf{Video}  & \textbf{Multiple} & & \\

\hline
 \multirow{24}{*}{\textbf{Image}} & \multirow{16}{*}{\textbf{Single}} &\multirow{1}{*}{\textbf{Structural Body Model}}  & \makecell[l]{$\bullet$ Spatial relationships of adjacent joints~\cite{14SP_TompsonJLB14,14SP_ChenY14,15SP_FanZLW15};\\ $\bullet$ Bi-directional tree-structured model~\cite{16SP_ChuOLW16}; \\ 
 {$\bullet$ Chain model~\cite{16SP_GkioxariTJ16};}
 \\$\bullet$ GAN-based pose discriminator~\cite{17SP_ChenSWLY17}; \\    $\bullet$ Human body compositional model~\cite{18SP_TangYW18}; \\ $\bullet$ Structured representation by GNN~\cite{19SP_abs-1901-01760}; \\ $\bullet$ Occlusion relational graphical model~\cite{17_ORGM}.}                                            \\ \cline{3-4} 
                                           &                                & \textbf{Multi-stage Pipeline} & \makecell[l]{ $\bullet$ Stacked hourglass~\cite{16SP_NewellYD16} and its variants~\cite{17SP_YangLOLW17,17SP_ChuYOMYW17,18SP_ChouCC18};\\ $\bullet$ CPM~\cite{SP16-CPM}  with intermediate  input  and  supervision.  }     \\ \cline{3-4} 
                                                 &                                & \multirow{1}{*}{\textbf{Pose Refinement}} & \makecell[l]{$\bullet$ Multi-model fusion~\cite{14SP_OuyangCW14} and Hybrid-Pose~\cite{20_MP_HybridRefine};\\ $\bullet$ Iterative update model~\cite{16SP_CarreiraAFM16,17SP_BelagiannisZ17}; \\ $\bullet$ Voting scheme~\cite{16SP_LifshitzFU16}; \\ $\bullet$ Coarse-to-fine hierarchical network~\cite{17SP_HuangGT17} and HCRN~\cite{19_ContextualRefine}; \\$\bullet$ Data-driven augmentation~\cite{18SP_FieraruKPS18,19MP_PoseFix}.}                                               \\ \cline{3-4} 
                                               &                                &\multirow{1}{*}{\textbf{Multi-task Learning}} & \makecell[l]{$\bullet$ Jointly 2D and 3D pose estimation~\cite{18SP_LuvizonPT18};\\ $\bullet$ Human parsing guided~\cite{18SP_NieFZY18}; \\ $\bullet$ Jointly train augmentation and pose estimation~\cite{18SP_PengTYFM18}.}   \\ \cline{3-4} 
                                           &                                & \multirow{1}{*}{\textbf{Efficiency Improvement}} &  \makecell[l]{$\bullet$ Multi-resolution and low computational cost~\cite{16SP_RafiLGK16}; \\$\bullet$ Binarized neural network~\cite{17SP_BulatT17a}; \\ $\bullet$ Hierarchical multi-scale residual architecture~\cite{17SP_BulatT17a}; \\ $\bullet$ Hourglass  using MobileNet~\cite{18SP_DebnathOYB18}; \\ $\bullet$ Pose distillation~\cite{19SP_ZhangZ019}. }  \\ \cline{2-4} 
                                                 & \multirow{7}{*}{\textbf{Multiple}}  &\multirow{1}{*}{ \textbf{Top-down}} & \makecell[l]{$\bullet$ Single stage model~\cite{17MP_papandreou2017towards,18MP_xiao2018simple}; \\ $\bullet$ Multi-task  (Whole body  pose   ZoomNet~\cite{20_Wholebody_topdown}, Mask-RCNN~\cite{17BU_he2017mask}, \\pose and parsing together~\cite{19MP_sanchez2019multi},~\cite{17MP_xia2017joint},and~\cite{19_LIP});
                                                 \\$\bullet$ Multi-stage/branch fusion (CPN~\cite{19MP_su2019multi}, MSPN~\cite{18MP_li2019rethinking}, RSN~\cite{20_MP_DelicateLearning},HRNet~\cite{19MP_HrnetXLW19}, Graph-PCNN~\cite{20_MP_Graph-PCNN}); \\$\bullet$ Complex case (RMPE~\cite{17MP_fang2017rmpe}, CrowdPose~\cite{19MP_li2019crowdpose},    OASNet~\cite{20_OcclusionSiamese}, ASDA~\cite{20_AdversarialAug}). } \\ \cline{3-4} 
                                                 &                                & \multirow{1}{*}{\textbf{Bottom-up}} & \makecell[l]{$\bullet$ Integer linear program for joint grouping (DeepCut~\cite{16BU_pishchulin2016deepcut},  DeeperCut~\cite{16BU_Insafutdinovdeepercut}); \\$\bullet$ Part Affinity Fields for 
                                                joint grouping (OpenPose~\cite{openpose},  PifPaf~\cite{19BU_kreiss2019pifpaf},  \\whole body OpenPose~\cite{19BU_hidalgo2019single}, and~\cite{2020BU_li2020simple}); 
                                                \\$\bullet$ Associative embedding for joint grouping~\cite{17BU_newell2017associative} and  HigherHRNet~\cite{cheng2020higherhrnet}; \\$\bullet$ Pose Partition Network~\cite{18BU_nie2018pose,19BU_nie2019single}; \\ $\bullet$ Multi-task (MultiPoseNet~\cite{18BU_kocabas2018multiposenet} and PersonLab~\cite{18BU_papandreou2018personlab}).  }                                          \\ \cline{1-4} 
                     \multirow{8}{*}{\textbf{Video}} & \multirow{1}{*}{\textbf{Single}} &    \multirow{1}{*}{\textbf{Temporal clues}} & \makecell[l]{$\bullet$  Insert multiple frames into channel layer~\cite{VOP14_PfisterSCZ14}; \\ $\bullet$ Along with action recognition (\cite{VOP15_NieXZ15} and~\cite{VOP17_IqbalGG17}); \\$\bullet$ Optical flow-based  model (Thin-Slicing~\cite{VOP17_Jiesong17} and \cite{VOP15_PfisterCZ15},~\cite{VOP16_CharlesPMHZ16});\\ $\bullet$ Sequence model (Chained Model~\cite{16SP_GkioxariTJ16}, LSTM Pose Machine~\cite{VOP18_LuoRWSPLPL18},  UniPose-LSTM~\cite{VOP20_artacho2020unipose}); \\$\bullet$ Dynamic Kernel Distillation~\cite{VOP19_NieLLZF19}.} \\ \cline{2-4} 
                                                  & \multirow{4}{*}{\textbf{Multiple}}  &  \multirow{1}{*}{\textbf{Top-down}} & \makecell[l]{$\bullet$ Clip-based spatio-temporal model (Detect-and-Track~\cite{VOP18_Detect-and-Track18} and~\cite{VOP20_wang2020combining});\\  $\bullet$  Optical flow-based FlowTrack~\cite{18MP_xiao2018simple} and PoseFlow~\cite{VOP18_PoseFlow}; \\
                                                  $\bullet$ Transformer-based keypoint tracker KeyTrack~\cite{KeyTrack}; \\ 
                                                  $\bullet$ Recovering missing detection (PGPT~\cite{PGPT} and~\cite{20_VOP_Self-supervised});  \\$\bullet$ Learnable similarity metric (POINet~\cite{POINet} and~\cite{20_VOP_TemporalMatching}.}     \\ \cline{3-4} 
                                                &                                &  \multirow{1}{*}{\textbf{Bottom-up}}&  \makecell[l]{$\bullet$ Graph partitioning-based model~\cite{VOP17_Posetrack17, VOP17_InsafutdinovAPT17}; \\ $\bullet$ Temporal Flow Fields-based model~\cite{VOP18_jointflow, JTA, VOP19_HwangLPK19, VOP19_RaajIHS19};\\ $\bullet$ Spatio-temporal associative embedding model KE-SIE~\cite{VOP19_JinLO019}.  }       \\ \hline
 \end{tabular}}
\end{table*}


\section{Monocular 2D Pose Estimation}
Monocular 2D pose estimation predicts the 2D locations of body keypoints in images or videos.  According to the input/output, the task can be divided into single person pose estimation and multi-person pose estimation in image-level or video-level.  Since the flexibility of the human body, 2D pose estimation has to deal with various postures, self-occlusion, and the interaction between body and scene. Especially, in multi-person scenes, the problems of crowd and occlusion further challenge the power of algorithms. In this section, we introduce the representative approaches according to the above categories and summarize them in Table~\ref{table_all_methods}. Additionally, we also give a brief introduction to the related tasks which use 2D pose estimation, such as person re-identification, action recognition, human-object interaction, human parsing,  etc.

\subsection{Single Person Pose Estimation}

As shown in Fig.~\ref{fig:outline}, the framework of typical single person pose estimation methods can be formulated as consisting of a pose encoder followed by a pose decoder. The pose encoder is a backbone to extract high-level features, while the pose decoder yields the 2D locations of keypoints in the regression-based manner or detection-based manner.

Most of the pose encoders are based on image classification networks, such as ResNet~\cite{resnet}, with a pre-trained model on a large-scale dataset such as ImageNet. Instead, few work designs the task-specific pose encoders. For example, the stacked hourglass network~\cite{16SP_NewellYD16} exploits the skip connection layer
to connect the mirror features with the same resolution. Furthermore, PoseNAS~\cite{20_MP_PoseNAS} exploits the Neural Architecture Search~\cite{DARTS} to find that the task-driven searchable feature extractor blocks. It directly searches a data-oriented
pose encoder with stacked searchable cells, which can provide an optimum feature extractor for the pose specific task.

Most of the recent works focus on the design of pose decoder, which pays more and more attention to explore the context information and the inherent characteristics of body structure. 
Toshev et al.~\cite{14SP_ToshevS14} propose DeepPose, which is one of the first human pose estimation methods based on deep convolutional neural networks (DCNNs). With a cascade of DCNN-based pose predictors, DeepPose formulates the keypoint estimation as a regression problem. It is different from previous traditional methods like manually designed graphical models~\cite{Dataset_FLIC,PCK} and part detectors~\cite{06PD_Ramanan06,09PD_AndrilukaRS09,13PD_LadickyTZ13}.
Iterative Error Feedback (IEF) network~\cite{16SP_CarreiraAFM16} exploits  
a self-correcting regression model. It is a kind of top-down feedback to progressively change the initial keypoint predictions.
Sun et al.~\cite{sun2017compositional} introduce the compositional pose regression, which
is body structure-aware.
The method in~\cite{18SP_LuvizonPT18} solves the regression-based keypoint prediction along with human action recognition in the multi-task manner.

Since the regression-based method directly maps the image to the coordinates of body joints, it is a non-linear problem and may fail for complex poses. 
Instead, the detection-based pose decoder generates heatmaps of keypoints instead of direct regression~\cite{14SP_TompsonJLB14}. As the detection-based pose decoders are widely used in many existed methods, we will introduce them according to their design categories as following.

\begin{figure*}[!t]
	\centering
	\captionsetup{justification=centering}
	\includegraphics[width=0.99\textwidth]{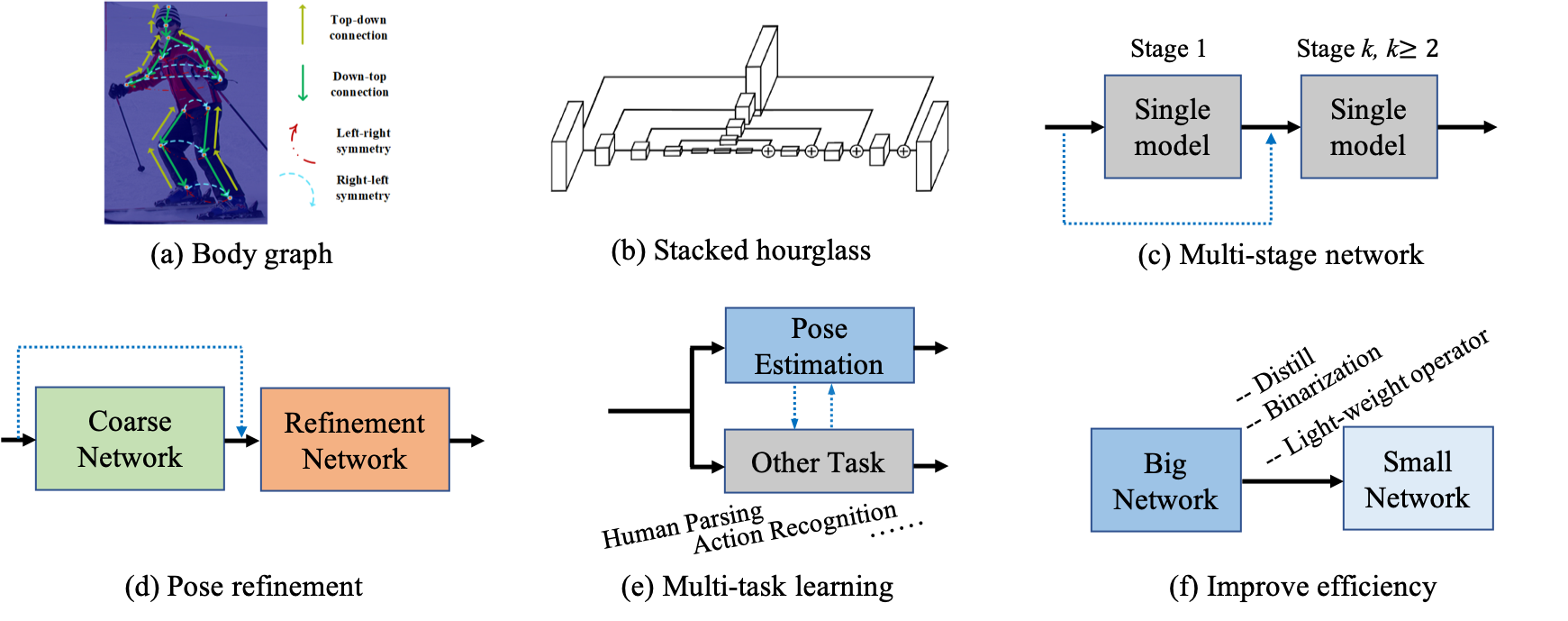}
	\caption{Illustration of six widely used paradigms for 2D single person pose estimation.}
	\label{fig:some_work_SPE}
    \end{figure*}

	\textbf{Structural Body Model}. Along with the DCNN-based feature representation for the whole body, graphical models are explored to describe the structural and local parts with the spatial relationship, as illustrated in Fig.~\ref{fig:some_work_SPE} (a). 
	Tompson et al.~\cite{14SP_TompsonJLB14} propose the convolutional network Part-Detector via a hybrid DCNN architecture. They formulate the distribution of spatial locations for body parts as an Markov Random Field-like model, which helps to remove the anatomically incorrect pose predictions.
   Similarly, Chen et al.~\cite{14SP_ChenY14} use DCNNs to learn conditional probabilities for the presence of body parts and their spatial relationships within image patches. 
   Different from those works that learn pair-wise relationship from the predicted score maps, Chu et al.~\cite{16SP_ChuOLW16} first investigate the relationship among parts at the feature level. The proposed end-to-end learning framework captures structural information among body joints by the learnable geometrical transform kernels and a bi-directional tree-structured model.
   Other than relying on any assumptions about the conditional distributions of joints, Gkioxari et al.~\cite{16SP_GkioxariTJ16} propose a chained sequence-to-sequence model to sequentially predict each body part based on all previously predicted body parts.
   Besides, to avoid biologically implausible pose predictions, the work in~\cite{17SP_ChenSWLY17} proposes a structure-aware network to implicitly exploit geometric constraint priors of the human body. It designs discriminators to distinguish the real poses from the fake ones by the conditional Generative Adversarial Networks (GANs).
   To further learn the compositionality of human body, Tang et al.~\cite{18SP_TangYW18} propose the deeply learned compositional model (DLCM) that has the bottom-up/top-down inference stages across multiple semantic levels. In the bottom-up stage, the higher-level parts are recursively estimated from their children, while in the top-down stage, the lower-level parts are recursively refined by their parents.
   Different from the previous approaches that use fully shared features for all body parts, Tang et al.~\cite{19SP_TangW19} proposes to learn specific features for related parts. Moreover, instead of the manually defined body structure relation, they propose a data-driven approach to group related parts based on the amount of information they shared. 
	Additionally, to deal with occlusion, ORGM~\cite{17_ORGM} proposes an occlusion relational graphical model to represent the self-occlusion and object-person occlusion simultaneously, which discriminatively encodes the interactions between human body parts and objects.

	\textbf{Multi-stage Pipeline}. It has been shown that multi-stage pipeline and multi-level feature fusion (illustrated in Fig.~\ref{fig:some_work_SPE} (c)) are useful for capturing the details of the human body. 
	One of the representative work is the stacked hourglass network~\cite{16SP_NewellYD16}, as shown in Fig.~\ref{fig:some_work_SPE} (b).  Each hourglass network consists of a symmetric distribution between bottom-up processing (from high resolutions to low resolutions) and top-down processing (from low resolutions to high resolutions). It uses a single pipeline with skip layers to preserve spatial information at each resolution. In conjunction with the intermediate supervision, the whole network consecutively stacks multiple hourglass modules together. It has been a solid baseline for its variants~\cite{17SP_YangLOLW17,17SP_ChuYOMYW17,18SP_ChouCC18} with various network design optimization. Among them, Yang et al.~\cite{17SP_YangLOLW17}  propose to insert the designed pyramid residual modules into the hourglass network, which can handle scale changes among human body parts.
	The work in~\cite{17SP_ChuYOMYW17}  designs the Hourglass Residual Units (HRUs) to increase the receptive field of the stacked hourglass network. Meanwhile, a multi-context attention mechanism is exploited to enable the representation of different granularity from local regions to global semantic consistent spaces.
	To exploit the structural information and multiple resolution features, the method in~\cite{18SP_KeCQL18} exploits the multi-scale supervision, multi-scare regression, and structure-aware loss on the stacked hourglass framework. 
   Besides stacked hourglass, another well-known multi-stage network Convolutional Pose Machine (CPM)~\cite{SP16-CPM} uses the intermediate input and supervision to learn implicit spatial models without an  explicit graphical model. Its sequential multi-stage convolutional architectures increasingly refine the prediction for keypoint locations.  

	\textbf{Pose Refinement}. Refinement for the network outputs can improve the final pose estimation performance.  Fig.~\ref{fig:some_work_SPE} (d) shows the framework of the common coarse-to-fine refinement pipeline. Ouyang et al.~\cite{14SP_OuyangCW14} build a multi-source deep model to extract non-linear representation from different information sources, including visual appearance score, appearance mixture type and deformation. Th representations of all information sources are fused for pose estimation. It can be viewed as the post-processing of pose estimation results. 
    The work in~\cite{16SP_CarreiraAFM16} uses an iterative update module to progressively make an incremental improvement to the pose estimation. Belagiannis et al.\cite{17SP_BelagiannisZ17} introduce a recurrent convolutional neural network to iteratively improve the performance.
	Lifshitz et al.~\cite{16SP_LifshitzFU16} propose a voting scheme for optimal pose configuration where each pixel in the image votes for the optimal position of each keypoint.
	Besides, there are some methods that use multi-branch networks for pose refinement. Huang et al.~\cite{17SP_HuangGT17} present a coarse-fine hierarchical network consisting of multiple branches. With multi-level supervision for the multi-resolution feature maps, multiple branches are unified to predict the final keypoints.
    HCRN~\cite{19_ContextualRefine} is a hierarchical contextual refinement network in which keypoints of different complexities are processed at different layers. HCRN is in a single-stage pipeline by exploiting the contextual refinement unit to transfer informative context from easy joints to difficult ones.
    Hybrid-Pose~\cite{20_MP_HybridRefine} adopts a two-branch Stacked Hourglass Networks, a Refinement Network (RNet) for pose refinement, and a Correction Network (CNet) for pose correction. RNet refines the keypoint locations in each hourglass stage horizontally. CNet guides the refinement and fuses the heatmaps in a hybrid manner.

Different from adding an extra network to the ahead coarse network for end-to-end training, the works in~\cite{18SP_FieraruKPS18} and~\cite{19MP_PoseFix} apply a similar refinement strategy  to take both the RGB images and the coarse predicted keypoints as input. Then the refinement network directly predicts a refined pose by jointly reasoning the input-output space. This kind of separate refinement network employs a data-driven augmentation for training and can be applied to any existing method.
	
	\textbf{Multi-task Learning}. As shown in Fig.~\ref{fig:some_work_SPE} (e), by exploiting complementary information from the related tasks, multi-task learning can provide extra cues for pose estimation. For example, Luvizon et al.~\cite{18SP_LuvizonPT18} propose a multi-task framework for jointly 2D/3D pose estimation and human action recognition from video sequences. 
	The method in~\cite{18SP_NieFZY18} uses a human part parsing learner to exploit the part segmentation information and provide complementary features to assist pose estimation. The adversarial data augmentation is exploited in~\cite{18SP_PengTYFM18} to address the limitation of random data augmentation during network training. It also designs a reward/penalty strategy for jointly training the augmentation network and the target (pose estimation) network.
	
	\textbf{Improving Efficiency}. Along with the development of model performance, how to improve the speed of a model has also attracted lots of attention. Fig.~\ref{fig:some_work_SPE} (f) shows the commonly used framework for improving model efficiency, including using light-weight operator, network binarization, model distillation, etc.
	RafiLGK et al.~\cite{16SP_RafiLGK16} propose a multi-resolution light-weight network
	that explores low computational requirements.
	The Binarized neural network is first exploited in~\cite{17SP_BulatT17a} to design a light-weight network with limited computational resources. Specifically, based on an exhaustive evaluation of various design choices, a hierarchical, parallel, and multi-scale residual architecture is proposed.
	The method in~\cite{18SP_DebnathOYB18} investigates the combination of MobileNets and the hourglass network to design a light-weight architecture.
	In addition, the work in~\cite{19SP_ZhangZ019} presents a pose distillation (FPD) model that trains a high-speed pose network based on the idea of knowledge distillation. 
	
    \begin{figure*}[!t]
	\centering
	\captionsetup{justification=centering}
	\includegraphics[width=0.99\textwidth]{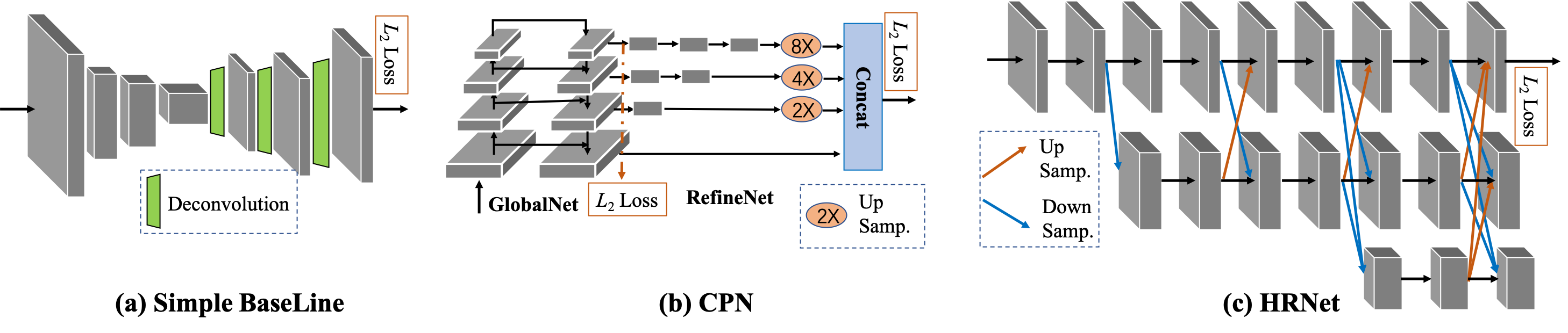}
	\caption{Three representative top-down 2D multi-person pose estimation networks: (a) Simple BaseLine~\cite{18MP_xiao2018simple}; (b) CPN~\cite{18MP_chen2018cascaded}; and (c) HRNet~\cite{19MP_HrnetXLW19}.}
	\label{fig:some_work_MPE_TP}
    \end{figure*}
	
	\subsection{Multi-person Pose Estimation}
Multi-person pose estimation needs to detect and locate the keypoints of all persons in an image, where the number of persons is unknown in advance. According to the processing paradigm, the representative methods can be sorted into two categories, i.e., top-down methods and bottom-up methods. The former is a two-stage pipeline that firstly detects all persons in an input image, then detects keypoints of each person in the detected bounding box. Differently, the bottom-up pipeline predicts all keypoints at once, then assigns these keypoints to different persons. We will introduce the representative CNN-based methods of these two categories. 

\subsubsection{Top-down Methods}

This kind of methods firstly detect and crop each person in the image. Then given a cropped image patch that only contains a single person, they use single-person pose estimation models followed by post-processing, such as pose Non-Maximum-Suppression (NMS)~\cite{17MP_papandreou2017towards}, to predict the final keypoint outputs of each person. Theoretically, the single person methods introduced in Section III.A can be applied after cropping the image patch. However, compared with the single person case, multi-person scenes have to deal with truncation, environmental occlusion, person-person occlusion, and small targets. Therefore, the representative top-down methods not only focus on designing networks by digging the potential of CNN and exploring rich context information fusion or exchange, but also pay attention to complex scenes. 

\textbf{Two Stage Pipeline}. Papandreou et al.~\cite{17MP_papandreou2017towards}  propose one of the first deep learning-based two-stage top-down pipeline, named G-RMI, which achieves the state-of-art results on the challenging COCO 2016 keypoints task. They use the Faster RCNN detector to detect each person, then exploit a fully convolutional ResNet~\cite{resnet} to jointly predict the keypoint's dense heatmaps and offsets. They also introduce the keypoint-based NMS instead of the box-level NMS to improve the keypoint confidence.  Furthermore, as in Fig.~\ref{fig:some_work_MPE_TP} (a), Xiao et al.~\cite{18MP_xiao2018simple} provide a simple and effective model that consists of a ResNet backbone and three deconvolution layers to increase the spatial resolution. It shows that a well-designed simple top-down model can achieve surprisingly effective. 

\textbf{Multi-task Learning}. 
By sharing features between related tasks of pose estimation, multi-task learning can provide better feature representations for pose estimation. For example, Mask-RCNN~\cite{17BU_he2017mask} can detect person bounding boxes, then crops the feature map of the corresponding proposal to predict human keypoints. Since human keypoints and human semantic parts are related and complementary,  many works~\cite{19MP_sanchez2019multi, 17MP_xia2017joint, 19_LIP} design multi-task networks to jointly predict the keypoints and segment the semantic parts. 
Besides, ZoomNet~\cite{20_Wholebody_topdown} unifies the human body pose estimator, hand/face detectors, and hand/face pose estimators into a single network. The network first localizes the body keypoints, then zooming in the hands/face regions to predict those keypoints with higher resolutions. It can handle the scale variance among different human parts. by  Moreover, to deal with the lack of the whole-body data, the COCO-WholeBody dataset is proposed by extending the COCO dataset with whole-body annotations.

\textbf{Multi-stage or Multi-branch Fusion}. Multi-stage or multi-branch fusion strategy is developed to break the bottleneck of a single model. The work in~\cite{18MP_chen2018cascaded} proposes a Cascade Pyramid Network (CPN), as shown in Fig.~\ref{fig:some_work_MPE_TP} (b), which consists of a global network and a refining network to progressively refine the keypoint prediction. It also proposes an online hard keypoints mining (OHKM) loss to deal with hard keypoints. CPN achieves the 1st place in the COCO 2017 keypoint challenge. The work in~\cite{19MP_su2019multi} improves CPN by introducing the channel shuffle module and the spatial channel-wise attention residual bottleneck to boost the original model. MSPN~\cite{18MP_li2019rethinking}, the winner of the COCO 2018 keypoint challenge, extends CPN in the multi-stage pipeline. It uses the global network of CPN as each single-stage module, fuses features from different stages by the cross-stage feature aggregation, and supervises the whole network via the coarse-to-fine loss functions. HRNet~\cite{19MP_HrnetXLW19}, shown in Fig.~\ref{fig:some_work_MPE_TP} (c), points out that the high-resolution representation is important for hard keypiont detection. HRNet maintains the high-resolution
representations through the whole network, and gradually adds high-to-low resolution sub-networks to form multi-resolution features. It has been a solid and superior model for pose estimation and many other computer vision tasks. Furthermore, to consider the keypoints' relationship and refine the rough predictions, Graph-PCNN~\cite{20_MP_Graph-PCNN} proposes a graph pose refinement module Via a model-agnostic two-stage framework. 
The work~\cite{20_MP_DelicateLearning} of the 1st place of COCO Keypoint Challenge 2019 utilizes a  multi-stage pipeline with Residual Steps Network (RSN) modules to aggregate intra-level features. With the delicate local representations obtained from RSN, a Pose Refine Machine (PRM) module is proposed to further balance the local/global representations and refine the output keypoints.
The resulting architecture establishes the new state of the art on the COCO dataset and MPII dataset.

\textbf{Dealing with Complex Scenes:}
In real-world applications, crowded, occlusion, and truncation scenes are unavoidable.
To remove the effect of inaccurate person detection, RMPE~\cite{17MP_fang2017rmpe} designs a symmetric spatial transformer network to detect every person, a parametric pose NMS to filter out the redundant pose, and a pose-guided human proposal generator to enhance the network capacity for multi-person pose estimation.
To tackle the problem in crowded scenes, Li et al.~\cite{19MP_li2019crowdpose} firstly obtain joint candidates in each cropped bounding box, then solve the joint association problem in a graph model. They also collect a crowded human pose estimation dataset named CrowdPose, and define the Crowd Index to measure the crowding level of an image.  
The work in~\cite{19MP_golda2019human} investigates the problem of pose estimation in crowded and occlusion surveillance scenes. It proposes to add an extra network branch to detect occluded keypoints. Besides, OASNet~\cite{20_OcclusionSiamese} exploits the Siamese network with an attention mechanism to remove the occlusion-aware ambiguities and reconstruct the occlusion-free features. To enlarge the training set for challenging cases, Bin et al.~\cite{20_AdversarialAug} propose to augment images by combing segmented body parts to simulate challenging examples. A generative network is utilized to dynamically adjust the augmentation parameters and produce the most confusing training samples. 


    \begin{figure*}[!t]
	\centering
	\captionsetup{justification=centering}
	\includegraphics[width=0.95\textwidth]{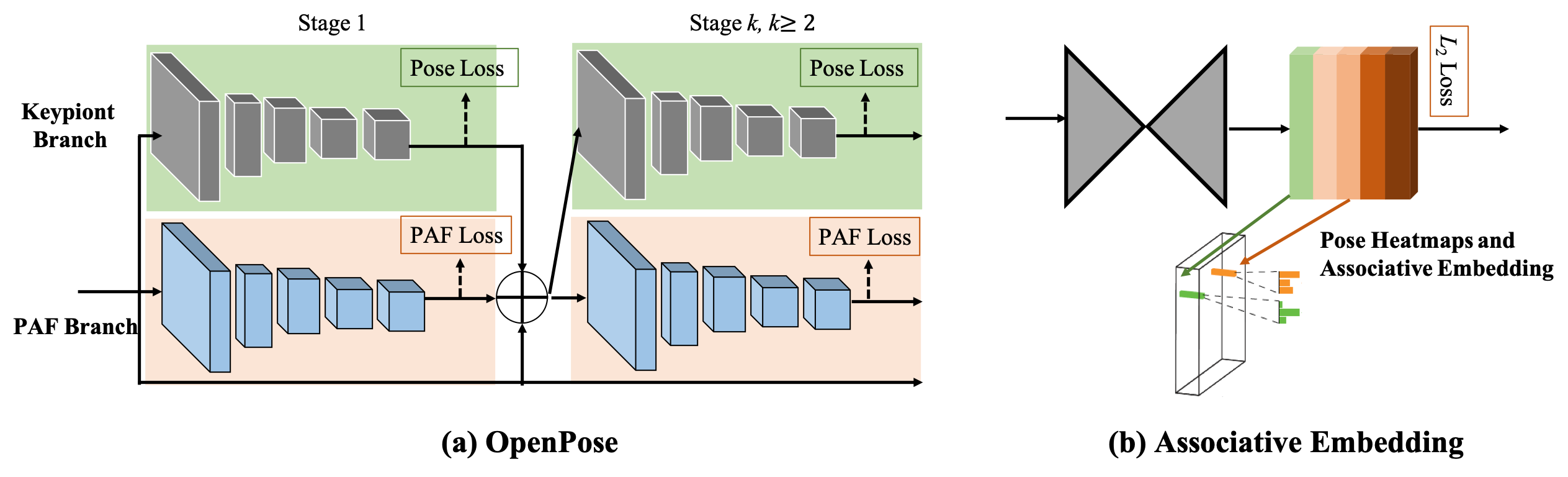}
	\caption{Two representative bottom-up 2D multi-person pose estimation methods: (a) OpenPose~\cite{openpose}; and (b) Associative Embedding~\cite{17BU_newell2017associative}.}
	\label{fig:some_work_MPE_BU}
    \end{figure*}

\subsubsection{Bottom-up Methods}

Different from top-down methods that rely on the human detector, bottom-up methods directly predict all keypoints in the image, then group keypoint candidates into each person. Besides the network design for more accurate keypoint detection, how to encode the connection information between keypoints is the core for grouping keypoints to different people. In the early stage, DeepCut~\cite{16BU_pishchulin2016deepcut} predicts all keypoints by Fast-RCNN~\cite{ren2015faster} and formulates the keypoint assignment problem as an integer linear programming (ILP).  DeeperCut~\cite{16BU_Insafutdinovdeepercut} improves DeepCut by introducing a stronger part detector. Besides, it proposes an image-conditioned pair-wise term that explores keypoint geometric and appearance constraints. However, ILP is still a time-consuming NP-hard problem. Differently, recent works exploit the learnable association strategies for keypoint grouping, such as Part Affinity Fields~\cite{openpose}, Associative Embedding~\cite{17BU_newell2017associative}, and their variants. We will further introduce them in detail.

\textbf{Part Affinity Fields}. 

The most popular bottom-up pose estimation method OpenPose~\cite{openpose} proposes to jointly learn keypoint locations and their association by the Part Affinity Fields (PAFs). As shown in Fig.~\ref{fig:representation}(c), PAF encodes the location and orientation of limbs by a set of 2D vector fields. The direction of PAF points from one part of the limb to the other. Then multi-person association performs the bipartite matching to associate keypoint candidates using PAFs. Via a two-branch and multi-stage architecture, as shown in Fig.~\ref{fig:some_work_MPE_BU}(a), OpenPose achieves real-time performance independent of the people number in the image. The idea of PAF is further explored by~\cite{19BU_kreiss2019pifpaf, 19BU_hidalgo2019single,2020BU_li2020simple}. In PifPaf net~\cite{19BU_kreiss2019pifpaf}, the PAF is used to associate body parts. In addition, the Part Intensity Field (PIF) is designed to localize body parts. PIF and PAF are jointly produced along with the utilization of Laplace loss to deal with the low-resolution and occluded scenes. Moreover, Hidalgo et al.~\cite{19BU_hidalgo2019single} propose the first single-network approach for whole-body multi-person pose estimation, which simultaneously locates the body, face, hands, and feet keypoints in an image. The work in~\cite{2020BU_li2020simple} designs a body part-aware PAF to encode the connection between keypoints, and improves the stacked hourglass network with attention mechanisms and a focal $L_2$ loss.

\textbf{Associative Embedding}. 
This associative embedding is a kind of detection and grouping method, which detects keypoints and group them into persons with embedding features or tags. Newell et al.~\cite{17BU_newell2017associative} propose to generates the keypoint heatmaps and their embedding tags for multi-person pose estimation. As shown in Fig.~\ref{fig:some_work_MPE_BU} (b), for each joint of the body, the network produces detection heatmaps and predicts associative embedding tags at the same time. They take the top detections for each joint, and match them to other detections which share the same embedding tag, to produce a final set of individual pose predictions. 
Associative Embedding is also used by HigherHRNet~\cite{cheng2020higherhrnet}, which learns scale-aware representations from high-resolution feature pyramids. Exploiting the aggregated features from HRNet~\cite{19MP_HrnetXLW19}, and the up-sampled higher-resolution features through a transposed convolution, HigherHRNet well handles the scale variation and achieves the new state-of-the-art for bottom-up pose estimation.
For other embedding-based keypoint assignment strategies, Nie et al.~\cite{18BU_nie2018pose} propose a Pose Partition Network (PPN) that uses the centroid embedding for all keypoint candidates. \cite{19BU_nie2019single} introduces the Structured Pose Representation (SPR). It exploits the root joints to indicate different people and encodes the positions of keypoints into their displacements w.r.t. the corresponding root.

\textbf{Multi-task Learning}. Besides pose estimation alone, MultiPoseNet~\cite{18BU_kocabas2018multiposenet} proposes a multi-task model that can jointly handle person detection, keypoint detection, and person segmentation. 
MultiPoseNet proposes a Pose Residual Network (PRN) to assign the predicted keypoints and the detected person bounding boxes by measuring their locations' similarity.
PersonLab~\cite{18BU_papandreou2018personlab} is also a multi-task network that jointly predicts keypoint heatmaps and person segmentation maps. 
In PersonLab, short-range offsets and mid-range pairwise offsets are used to group human keypoints. Meanwhile, long-range offsets and the human pose detections are used to distinguish person segmentation masks.

\subsection{2D Pose Estimation in Videos}
With the development of image-based pose estimation methods, 2D pose estimation in videos has also been boosted by DCNNs. Different from image-based pose estimation, video pose estimation must consider  the temporal relation across frames to remove the motion blur and geometric inconsistency. Therefore, directly applying the existing image-based pose estimation methods on videos may produce sub-optimal results. In this part, we can classify the single/multi person(s) pose estimation methods in videos according to how do they exploit the spatio-temporal information. 

\subsubsection{Single Person Pose Estimation in Videos}

For single person pose estimation in videos, most works explore to propagate temporal clues across frames for refining the single frame pose results. As one of the first works for deep learning based pose estimation in videos~\cite{VOP14_PfisterSCZ14}, Pfister et al. exploits temporal information by inserting multiple frames into the data colour channels to replace the three-channel RGB image input in an image-based network. The following works further explore the temporal propagation in four main categories, i.e., action recognition task for mutual learning,  temporal propagation by optical flow-based model,  sequence model, and distillation model.

\textbf{Pose Estimation with Action Recognition}. Nie et al.~\cite{VOP15_NieXZ15} propose a spatial-temporal And-Or Graph (ST-AOG) model to combine video pose estimation with action recognition. By adding additional activity recognition branches based on optical flow and appearance features, the two tasks mutually benefit from each other. Furthermore, instead of an additional action recognition branch, the work in~\cite{VOP17_IqbalGG17} shows that the action  prediction can be achieved by incorporating activity priors using an action conditioned pictorial structure model. 

\textbf{Optical Flow-based Feature Propagation}. Pfister et al.~\cite{VOP15_PfisterCZ15} propose SpatialNet that temporally warps the neighboring frame heatmaps to the current frame through optical flow. Afater that, they also exploit a parametric pooling layer to combine the aligned heatmaps into a pooled confidence heatmap. The work in~\cite{VOP16_CharlesPMHZ16} proposes a personalized ConvNet pose estimator, which can propagate high-quality automatic pose annotations throughout the video by spatial image matching and optical flow propagation. The propagated new annotations are used to fine-tune the generic ConvNet pose estimator. In Thin-Slicing~\cite{VOP17_Jiesong17}, Song et al. propose to propagate keypoints by computing dense optical flow between neighboring frames. They conduct a flow-based warping layer to align previous heatmaps to the current frame, followed by a spatio-temporal inference layer. The spatial-temporal propagation utilizes the iterative message passing of the pose configuration graph with both spatial and temporal relationships edges. 

\textbf{Sequence Model-based Feature Propagation}. In the Chained Model~\cite{16SP_GkioxariTJ16}, Gkioxari1 et al. adopt the sequence-to-sequence recurrent model to solve the structured pose prediction in videos. In the recurrent model, the prediction of each body keypoint relies on all previously predicted keypoints. In LSTM Pose Machine~\cite{VOP18_LuoRWSPLPL18}, Luo et al. explore to capture temporal dependency in videos by memory augmented LSTM framework.  Given a frame, the Encoder-RNN-Decoder pipelines~\cite{16SP_GkioxariTJ16, VOP18_LuoRWSPLPL18} firstly learn high-level image representations by an encoder, then propagate temporal information and produce hidden states by RNN units. They finally predict keypoints of the current frame by a decoder which takes hidden states as input.  
A similar concept is adopted by UniPose-LSTM~\cite{VOP20_artacho2020unipose}, which exploits an LSTM module to propagate previous heatmaps generated by the multi-resolution Atrous Spatial Pooling architecture in the UniPose network.

\textbf{Distillation Model}. Different from optical flow or sequential RNN that requires a large network, Nie et al.~\cite{VOP19_NieLLZF19} introduce a Dynamic Kernel Distillation (DKD) model to transfer pose knowledge in a one-shot feed-forward manner. Specifically, the small network DKD exploits a temporally adversarial training strategy via a discriminator to generate temporally coherent pose kernels and predict keypoints in videos.

\subsubsection{Multi-Person Pose Estimation and Tracking in Videos}

Multi-person pose estimation in videos has to deal with identifying and locating keypoints in the case of multi-person occlusion, motion blur, pose and appearance variations. To exploit temporal clues, multi-person pose estimation is usually associated with articulated tracking.  PoesTrack dataset~\cite{VOP17_Posetrack17, VOP18_Posetrack18}  is the first large-scale and in-the-wild multi-person dataset for pose estimation and tracking, which contains hundreds of videos with more than 150K poses.  
With the  boom of datasets, the mainstream solutions for this task can be sorted into top-down methods and bottom-up methods.

\textbf{Top-down Methods}. The top-down methods follow the tracking-by-detection paradigm. They first detect persons and keypoints in each frame, then propagate the bounding boxes or keypoints across frames. For example, 3D Mask R-CNN is proposed in Detect-and-Track~\cite{VOP18_Detect-and-Track18} to employ fully 3D convolutional networks to detect keypoints of each person in a video clip. Then a keypoint tracker is used to link the predictions by comparing the distances of the detected bounding boxes. Moreover, the clip-based tracker is further exploited in~\cite{VOP20_wang2020combining}, which extends HRNet via carefully designed 3D convolutional layers to learn the temporal correspondence between keypoints. Next, a spatio-temporal merging procedure is designed to estimate the optimal keypoint outputs by spatial and temporal smoothing. By combining the clip tracking network and the merging procedure, the model in~\cite{VOP20_wang2020combining} can well handle failure detections in complex scenes with severe occlusion and highly entangled people. 

Different from clip-based detection, the work in~\cite{18MP_xiao2018simple} independently builds on the single-frame detection and exploits the optical flow-based temporal pose similarity to associate keypoints across different frames. To cope with missing detection in a single frame, PGPT~\cite{PGPT} proposes to combine an image-based detector with an online person location predictor to compensate for the missing bounding boxes. Meanwhile, PGPT introduces a hierarchical pose-guided graph convolutional network that exploits the human structural relations to boost the person representation and data association. For efficient data association, the work in POINet~\cite{POINet} explores a pose-guided ovonic insight network to learn feature extraction, similarity metric, and identity assignment in an unified end-to-end network. In another work for learnable similarity metric~\cite{20_VOP_TemporalMatching}, temporal keypoint matching and keypoint refinement are designed for keypoints association and pose correction, respectively. They are both learnable and incorporated into a pose estimation network.
The work in~\cite{20_VOP_Self-supervised} proposes the self-supervised keypoint correspondences that cannot only recover missing pose detections, but also associate detected and recovered poses across frames.
Different from using high-dimensional image representation to track people, KeyTrack~\cite{KeyTrack} presents a transformer-based tracker that only relies on $15$ keypoints. The Transformer-based network exploits a binary classification  to predict whether one pose temporally follows another. 

\textbf{Bottom-up Methods}. This kind of methods commonly use the single frame pose estimation to predict all keypoints in each frame, then assign keypoints across frames in a spatio-temporal optimization manner. For example, the graph partitioning based methods in~\cite{VOP17_Posetrack17, VOP17_InsafutdinovAPT17,VOP18_Posetrack18} firstly extend the image-level bottom-up multi-person pose estimation~\cite{16BU_pishchulin2016deepcut, openpose}. Then they build a spatio-temporal graph that connects keypoint candidates spatially as well as temporally, which can be formulated into an linear programming problem. However, the spatial and temporal graph partitioning usually leads to heavy computation and non-online solution. Diffrerently, PoseFlow~\cite{VOP18_PoseFlow} exploits a pose flow that measures the pose distance in different frames to track the same person. 
Inspired by the  spatial Part Affinity Field in OpenPose~\cite{openpose}, the works in~\cite{VOP18_jointflow, JTA, VOP19_HwangLPK19, VOP19_RaajIHS19} exploit Temporal Flow Fields to indicate the propagating direction of keypoints in different frames.
Associative Embedding~\cite{17BU_newell2017associative}, which is used in the image-based bottom-up keypoint association strategy, is also extended in~\cite{VOP19_JinLO019} to build the spatio-temporal embedding. It associates keypoints with embedding features for temporal consistency.

\textbf{ \textit{In summary}}, the development of 2D pose estimation has been greatly boosted by the design of CNN-based networks with the aid of body part structural relationship, multi-stage pipeline, multi-level feature fusion, pose refinement, multi-task learning, and efficiency-aware design. In the multi-person case, an excellent top-down method relies on an accurate detection network and a solid single person pose estimation network. As for the bottom-up methods, the most important part is how to group the detected keypoints to different people.  In the video-level pose estimation and tracking, the representative researches focuse on how to efficiently propagate the spatio-temporal information to guarantee the consistency and smoothness of the predictions.

\subsection{2D Pose Estimation for Other Tasks} 

\textbf{Person Re-Identification: }
For the task of person re-identification (ReID), pose variations may bring significant change to the appearance of a person. The detailed part-based information is important to discriminate individuals, especially for the occluded person ReID and partial person ReID. For example, Su et al.~\cite{8237689} propose a Pose-driven model to learn improved feature extraction and matching models, where the response maps of body joints can alleviate the pose variations. Besides, Zhao et al.~\cite{8099586} exploits the body structure information to align body region features. Different semantic regions can be merged with a competitive scheme. To correct the pose variations caused by camera views and person motions, Zhang et al.~\cite{8693885} employ pose estimation to the body part discovery and affine projection. To deal with the misalignment and occlusion problem of ReID, Xu et al.~\cite{8578324} propose a confidence-map-based solution instead of the previous RoI-based solution, which can precisely exclude background clutter and adjacent part features. Furthermore, Miao et al.~\cite{miao2019pose} encode the position information of body parts, which helps to disentangle the useful information from the occlusion noise. Gao et al.~\cite{Gao_2020_CVPR} design a pose-guided visible part matching framework, which can jointly learn the discriminative pose-guided attention and part visibility in a self-supervised way.

\textbf{Action Recognition: }
Recognition of human actions is an important task in the understanding of dynamic scenes. The detection and alignment of human poses are proved to be crucial to capture discriminative information of human actions. For example, Cheron et al.~\cite{7410725} introduce pose estimation to extract the appearance and flow information at characteristic positions. This work points out the importance of a representation derived from the human pose. Furthermore, Nie et al.~\cite{VOP15_NieXZ15} propose a framework to integrate action recognition and pose estimation from video, where a spatial-temporal And-Or Graph model is employed for representing action and poses. Besides, Luvizon et al.~\cite{18SP_LuvizonPT18} propose a multi-task framework that jointly handles 2D/3D pose estimation and action recognition from video sequences. Differently, Du et al.~\cite{8237664} propose a pose-attention mechanism to adaptively learn pose-related features at every time-step action prediction of RNNs. With the development of 2D pose estimation, recent researches~\cite{du2015hierarchical,li2019actional,shi2020skeleton,yan2018spatial} tend to screen out the pose estimation task in action recognition and focus on skeleton-based action recognition directly with structured positions of human keypoints.

\textbf{Human-Object Interactions: }
Recognizing Human-Object Interactions (HOI) is an important research problem, which has massive applications in image understanding and robotics. Pose information is important to infer the interactions between the detected human and object. For example, Fang et al.~\cite{Fang_2018_ECCV}  propose a pairwise body-part attention model to focus on crucial parts and their correlations for HOI recognition. In the model, a pose estimator is employed to detect human keypoints and then body parts. Li et al.~\cite{Li_2019_CVPR} further combine visual appearance, spatial location, and human pose information for interactiveness discrimination. They employ pose estimation to predict the 17 keypoints to build spatial-pose information. Wan et al.~\cite{Wan_2019_ICCV} train CPN as the pose estimator, and utilize the estimated human pose to capture global spatial configuration as the guidance to extract local features at the semantic part level. In the 3D HOI task, Li et al.~\cite{Li_2020_CVPR} first obtain the 2D human pose, and then estimate 3D human body to construct the 3D spatial configuration volume.

\textbf{Human Parsing: }
Human parsing aims at segmenting a human image into different fine-grained semantic parts, which serves as the basis for many high-level applications, e.g., human behavior analysis, person re-identification, and video surveillance. The human pose can offer structure information for body part segmentation and labeling. For example, Dong et al.\cite{Dong_2014_CVPR} present a unified framework for simultaneous human parsing and pose estimation. They verify that the mutually complementary nature of the two tasks can boost the performance of each other.
Similarly, Nie et al.~\cite{nie2018mutual} present a novel mutual learning solution for joint human parsing and pose estimation. Besdies, Liang et al.~\cite{19_LIP} propose a joint human parsing and pose estimation network to explore efficient context modeling, which can simultaneously predict parsing and pose with high quality. Zhang et al.~\cite{Zhang_2020_CVPR} propose that human body edge and pose are two beneficial factors to human parsing. They design a superior way of leveraging the pivotal contextual cues provided by edges and poses for human parsing.

\begin{center}
\begin{table*}[]
 \setlength\tabcolsep{8pt}
\caption{Representative deep learning-based methods for monocular 3D pose estimation.}\label{table_all_methods_3d}
\footnotesize
\centering
\begin{tabular}{c|c|c|l}
\hline
 \textbf{Model Type}        &      \textbf{Input/Output Type}                       &              \textbf{Main Idea}                                                        &    \textbf{Methods}\\ 
\hline
 {\multirow{25}{*}{\textbf{Skeleton-based}}}        &  \multirow{19}{*}{\textbf{Single person}} &\multirow{1}{*}{\textbf{Heatmap-based}} & 
 \makecell[l]{ $\bullet$ Coarse-to-fine~\cite{pavlakos2017coarse};\\
 $\bullet$ VNect~\cite{mehta2017vnect};\\
 $\bullet$ Intergral heatmap regression IHP~\cite{sun2018integral}.\\}\\ \cline{3-4} 

&  & \multirow{1}{*}{\textbf{Lifting 2D pose to 3D}} & \makecell[l]{$\bullet$ Simple-3D~\cite{martinez2017simple};\\
$\bullet$ 2D-to-3D lifting via matching~\cite{chen20173d};\\
$\bullet$ Cycle consistency between 2D/3D~\cite{tome2017lifting,chen2019unsupervised}.\\
}\\ \cline{3-4}
&  & \makecell[l] {\textbf{Fusing image features with keypoints} } & \makecell[l]{$\bullet$ Global and local integration~\cite{nie2017monocular};\\
 $\bullet$ SemGCN~\cite{zhao2019semantic} and Liu et al.~\cite{liu2019feature}.\\
}\\ \cline{3-4}
                    
 &  &   \multirow{1}{*}{\textbf{Solving data lacking}}   & \makecell[l]{  
 $\bullet$ Depth ordering of keypoints~\cite{pavlakos2018ordinal,zhou2019hemlets};\\
 $\bullet$ Multi-view consistency~\cite{Rhodin_2018_CVPR,Rhodin_2018_ECCV,Kocabas_2019_CVPR,MV_2020_CVPR,Mitra_2020_CVPR};\\
 $\bullet$ Learning from synthetics~\cite{rogez2016mocap,chen2016synthesizing,surreal,kundu2020partpose}.\\
 } \\\cline{3-4}
 
 &  & \multirow{1}{*}{\textbf{Solving inherent ambiguity}} &
 \makecell[l]{ $\bullet$ Body Geometr prior~\cite{fang2018learning} ;\\
   $\bullet$ Temporal smoothness~\cite{lin2017recurrent, rayat2018exploiting,lee2018propagating,3dvideopose,Cheng_2019_ICCV};\\
   $\bullet$ Rationality/View supervision~\cite{sharma2019monocular,Kiciroglu_2020_CVPR}; \\
   $\bullet$ Hierarchy bone representation~\cite{DKA_Xu_2020_CVPR,ETD_2020_CVPR}. 
   } \\\cline{2-4} 
                   
&  \multirow{5}{*}{\textbf{Multi-person}}&  \multirow{1}{*}{\textbf{Top-down methods}}  & \makecell[l]{
$\bullet$ LCR-Net++~\cite{rogez2019lcr};\\
$\bullet$ 3DMPPE~\cite{moon2019camera}.\\} \\ \cline{3-4}     

&  &  \multirow{1}{*}{\textbf{Bottom-up methods}}  & \makecell[l]{
$\bullet$ Grouping keypoints to skeleton~\cite{zanfir2018deep};\\
$\bullet$ Occlusion-robust pose-maps~\cite{mehta2018single};\\
$\bullet$ Compressing volumetric heatmaps~\cite{fabbri2020compressed};\\
$\bullet$ PandaNet~\cite{PandaNet_Benzine_2020_CVPR};\\
$\bullet$ Depth estimation~\cite{zhen2020smap,wang2020hmor}.\\}      \\ \cline{1-4} 

 {\multirow{6}{*} {\textbf{Mesh-based}}}   & \multirow{6}{*} {\textbf{Single person}} &    \multirow{1}{*}{\textbf{Solving data lacking}} & \makecell[l]{
 $\bullet$ GAN-based rationality supervision~\cite{hmr,kocabas2019vibe};\\
 $\bullet$ Temporal dynamics~\cite{kanazawa2018learning,sun2019dsd-satn};\\
 $\bullet$ Appearance consistency~\cite{Xu_2019_ICCV,pavlakos2019texturepose};\\
 $\bullet$ SMPL optimization in the loop SPIN~\cite{kolotouros2019spin}.\\}\\ \cline{3-4}
 
 && \multirow{1}{*}{\textbf{Proper representations}} &   \makecell[l]{  
 $\bullet$ Graph representation GraphCMR~\cite{Kolotouros_2019_CVPR};\\
 $\bullet$ Voting results HoloPose~\cite{Guler_2019_CVPR};\\
 $\bullet$ Skeleton-disentangled representation~\cite{sun2019dsd-satn}); \\
 $\bullet$ Multi-body-part 3D mesh~\cite{smplh,monocualar,smplx}.\\
 }\\ \cline{2-4}
 
 &  \multirow{2}{*}{\textbf{Multi-person}}&  \multirow{1}{*}{\textbf{Multi-stage methods}}  & \makecell[l]{$\bullet$ Scene constraints~\cite{zanfir2018monocular};\\ $\bullet$ CRMH~\cite{Jiang_2020_CVPR}. } \\ \cline{3-4}     

&  &  \multirow{1}{*}{\textbf{Single-shot methods}}  & \makecell[l]{$\bullet$ CenterHMR~\cite{CenterHMR} . }     
                     \\ \hline
\end{tabular}
\end{table*} 
\end{center}

\section{Monocular 3D Pose Estimation}

Monocular 3D pose estimation can be divided into skeleton-based 3D pose estimation and mesh-based 3D pose estimation according to the output representation. The former predicts the 3D locations of the body joints, while the latter outputs 3D body mesh depending on a human mesh topology or a statistical 3D body model. Compared with the 2D pose estimation, estimating 3D pose from monocular 2D images is much more challenging. Besides all the challenges in the 2D part, monocular 3D pose estimation also suffers from the lack of in-the-wild 3D data and the inherent 2D-to-3D ambiguity. 

The first big challenge is lacking sufficient in-the-wild data with accurate 3D annotations. Most existed 3D pose datasets are not diversity enough. It is hard and expensive to precisely capture 3D pose annotations for 2D images, especially in outdoor conditions. Existing 3D pose datasets are often biased towards specific environments (e.g. indoor) with limited actions. For example, the well-known 3D pose dataset, Human3.6M~\cite{h36m}, only contains 11 actors performing 15 activities. In contrast, 2D pose data is easy to be collected, which contains much richer poses and environments. Therefore, 2D pose datasets~\cite{Dataset_COCO,Dataset_MPII,Dataset_LSP} are often used to improve the generalization of 3D algorithms. For example, most existing 3D pose estimation methods adopt 2D pose as an intermediate representation~\cite{sun2019dsd-satn,zhao2019semantic} or even network input~\cite{martinez2017simple,3dvideopose} for reducing the difficulties. Besides, many methods propose un-supervised~\cite{Rhodin_2018_ECCV,chen2019unsupervised} or weak-supervised~\cite{Rhodin_2018_CVPR,Mitra_2020_CVPR} frameworks to alleviate the dependence of fully supervised methods on datasets. 
  
    
Furthermore, losing the depth information may cause the inherent 2D-to-3D ambiguity problem for 3D pose estimation. Especially for the two stage-based 3D pose estimation methods, they first try to estimate 2D pose from images, then lift 2D pose to 3D. The inherent ambiguity problem of these methods are even serious because multiple 3D poses can map to the same 2D keypoints. Many methods attempt to tackle this problem by using various prior information, such as geometric prior knowledge~\cite{pavlakos2018ordinal,fang2018learning}, statistical model~\cite{smpl,hmr}, and temporal smoothness~\cite{tpnet,kocabas2019vibe}. 

In this section, according to the way of pose representation, we classify the representative 3D pose estimation methods into the 1) skeleton-based paradigm and 2)  mesh-based paradigm. Additionally, we also introduce ideas to solve the lack of in-the-wild 3D data, inherent 2D-to-3D ambiguity, and the case of the multi-person scene. We summarize all the representative methods in Table~\ref{table_all_methods_3d}.

	\begin{figure*}[!t]
	\centering
	\includegraphics[width=0.95\textwidth]{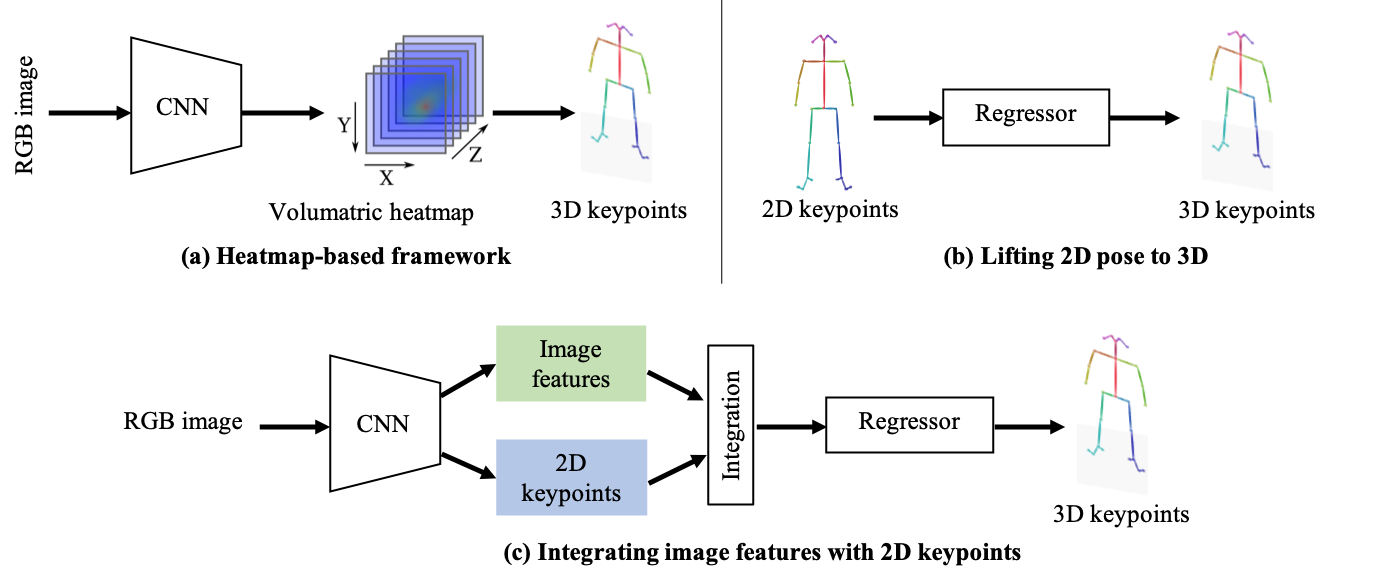}
	\caption{Representative frameworks of monocular 3D human pose estimation. }
	\label{fig:3d_framework}
    \end{figure*}

    \subsection{Skeleton-based 3D Pose Estimation}
	
	As shown in Fig.~\ref{fig:3d_framework}, according to the representation, three kinds of frameworks are generally adopted by the prevailing methods for single person 3D pose estimation, i.e., methods based on 1) volumetric heatmap, 2) lifting 2D pose to 3D pose, and 3) fusing image features with the 2D pose. Furthermore, because of the significance, we also summarize methods to solve two common challenging cases in 3D pose estimation, i.e., 4) the lack of 3D data and 5) 2D-3D inherent ambiguity. 
    
    
  \subsubsection{Heatmap-based Methods}
Different from the regression-based method~\cite{DconvMP} that directly predicts the 3D location of each keypoint, heatmap-based methods represent each 3D keypoint as a 3D Gaussian distribution in the heatmap. 
Then in post-processing, we can easily parse the 3D keypoint coordinates from the estimated volumetric heatmap via acquiring the local maximum. Therefore, as shown in in Fig.~\ref{fig:3d_framework} (a), the heatmap-based methods are designed to directly estimate the volumetric heatmap from the monocular images through an end-to-end framework. Following this pipeline, Coarse-to-Fine (C2F)~\cite{pavlakos2017coarse} network is proposed to stack the hourglass networks~\cite{16SP_NewellYD16}, and progressively extend the volumetric direction of the predicted heatmap for fine-gained results. Besides, VNect~\cite{mehta2017vnect} is a real-time monocular 3D pose estimation method. It makes the coherent kinematic skeleton fitting in post-processing to yield temporally stable pose results based on a coherent kinematic skeleton. To simplify the post-processing, Integral Human Pose (IHP)~\cite{sun2018integral} proposes an integral operation to directly convert the heatmaps into keypoint coordinates in a differentiable manner during forward inference. It builds a bridge between the heatmap-based and regression-based methods.

\subsubsection{Lifting 2D Pose to 3D}
Benefited from the robust 2D pose estimation methods, as shown in Fig.~\ref{fig:3d_framework} (b), many methods focus on lifting 2D pose to 3D via a simple regressor. The entire pipeline can be divided into two parts: 1) estimating 2D pose from monocular images, and 2) lifting the estimated 2D poses to 3D. In this way, they can build the 3D methods based on the 2D pose estimation. The model complexity is much lower, while the generalization can be improved. For example, Martinez et al.~\cite{martinez2017simple} propose Simple-3D, which is a well-known simple baseline method to estimate the depth of each keypoint from the 2D pose. It only contains two fully connected blocks while achieves good performance on the related benchmarks. Furthermore, Chen et al.~\cite{chen20173d} solve the 2D-to-3D keypoint lifting problem via pose matching. To enrich the matching library, they generate a lot of 2D-3D pose pairs via projecting the 3D pose back to the 2D image plane using a random camera. Then they get a large 2D-3D pose pairs library. Given a 2D image, they just need to predict the 2D pose, and search the most similar 2D-3D pose pairs from the library. The paired 3D pose is selected as the 3D pose estimation results. Differently, Tome et al.~\cite{tome2017lifting} develop a multi-stage convolution network to recurrently optimize the estimated 3D pose. They refine the intermediate 2D pose using the back projection of the estimated 3D pose, which progressively increases the accuracy of both 2D and 3D pose estimation. Besides, Chen et al.~\cite{chen2019unsupervised}  take advantage of the cycle consistency between 2D pose input and the 2D projection of the estimated 3D pose to learn the 2D-to-3D lifting function in an unsupervised manner.
Without any 3D annotations, the cycle consistency could not help the model to properly learn the depth. The model may converge to a local minimum with a constant depth. 
To tackle it, they take advantage of a geometric prior that 2D projections of the same 3D pose from different views should lift to the same 3D pose.
Specifically, the lifted 3D pose is firstly projected back to 2D image from a random view to supervise the depth-wise rationality using 2D pose available in a generative adversarial manner.
Then they lift the randomly projected 2D pose back to 3D and supervise the difference between its 2D projection from the original view and the original 2D pose input. 
In this way, the model get properly trained without any 3D annotations.

   
\subsubsection{Integrating Image Features with 2D Keypoints}
As shown in Fig.~\ref{fig:3d_framework} (c), some methods attempt to integrate the heatmap-based and lifting 2D pose to 3D frameworks together. On one hand, 2D keypoints provide limited information of the human body, which might make 2D-to-3D lifting ambiguous in challenging scenes. On the other hand, image features provide more contextual information, which are helpful to determine the accurate 3D pose. In this manner, Nie et al.~\cite{nie2017monocular} propose to integrate the local image texture features of keypoints into the global skeleton of the 2D pose. Then a two-level hierarchy of LSTM network is developed to model the global and local features progressively. Furthermore, SemGCN~\cite{zhao2019semantic} and Liu et al.~\cite{liu2019feature} both extract joint-level image features, and integrate them with keypoint coordinates to form multiple graph nodes. Next, GCN and LSTM are used to dig into the relationships between these nodes with the help of image features.

\subsubsection{Solving the Data Deficiency Problem }
Most 3D pose datasets capture very limited activities of a few actors in indoor environments. Compared with 2D pose datasets, 3D pose datasets are poor in the diversity of pose and environment. To handle this problem, many approaches attempt to train the model in an unsupervised or weak-supervised manner.
For example, some approaches propose to use weak but cheap labels for supervision. For example, Pavlakos et al.~\cite{pavlakos2018ordinal} take advantage of the weak ordinal depth relations between keypoints for supervision. Experiments show that compared with the direct supervision using ground truth 3D pose annotations, the ordinal supervision can also achieve comparative performance. Similarly, Hemlets~\cite{zhou2019hemlets} encodes the explicit depth ordering of adjacent keypoints via a heatmap triplet loss as ground truth. 

	Except for the body structure prior, many approaches propose to use multi-view consistency for supervision.  
    However, only considering the multi-view consistency will lead to a degenerated solution~\cite{Rhodin_2018_CVPR}. The model may be trapped into a local minimum and produce similar zero poses for different inputs. To address this problem, Rhodin et al.~\cite{Rhodin_2018_CVPR} propose to use a small amount of labeled data to avoid the local minimum and correct the predictions. Differently, Rhodin et al.~\cite{Rhodin_2018_ECCV} propose  to use sequential images to provide temporal consistency prior for body representation learning. EpipolarPose~\cite{Kocabas_2019_CVPR} takes advantage of the multi-view 2D poses for generating 3D pose annotations via the epipolar geometry. In this way, the entire framework can be trained in a self-supervised manner. Umar et al.~\cite{MV_2020_CVPR} tackle the degeneration trap via a novel alignment-based object function without requiring extrinsic camera calibration. They train the model using unlabelled multi-view images and 2D pose datasets. The 2D poses are only used to train the 2D pose backbone. The multi-view images are exploited for the corresponding 3D pose estimation. Multi-view 3D pose results are normalized and then aligned using Procrustes analysis for the consistency. In this process, the 3D poses are transformed to the same scale for rigid alignment. Mitra et al.~\cite{Mitra_2020_CVPR} propose the Multiview-Consistent Semi-Supervised Learning (MCSS) framework, which try to learn the view-invariant pose embedding in a semi-supervised manner. On one hand, the model is trained to estimate the view-invariant 3D poses by making the left pelvis bone parallel to the $XZ$ plane. On the other hand, temporal relation-based hard-negative mining is applied to gain consistent pose embedding in different views.

	Additionally, some methods propose to learn the model from data synthesis. 
	In general, there are two kinds of synthetic data pipelines: 1) 2D image stitching pipeline, and 2) 3D model projection pipeline. For the first framework, Rogez et al.~\cite{rogez2016mocap} attempt to generate the 2D image of 3D poses brought from 3D motion capture (MoCap) datasets. They first select the image patch whose 2D pose matches a part of the projected 3D pose. With kinematical constraints, local image patches are stitched up to form the complete 2D image of a 3D pose. 
    Differently, Chen et al.~\cite{chen2016synthesizing} and Varol et al.~\cite{surreal} follow the 3D modal projection pipeline. They project the textured statistical human body model onto the 2D in-the-wild background images for data generation, e.g., the Shape Completion and Animation of People (SCAPE)~\cite{scape} and SMPL~\cite{smpl}. In this way, the complete 3D annotations of a 2D in-the-wild person image is available. The annotations contain not only the 3D body pose, shape, and texture, but also the camera and light parameters. In summary, the 2D image stitching pipeline has the potential to generate more realistic person images, while the 3D model projection pipeline can obtain more comprehensive 3D annotations. Besides, PGP-human~\cite{kundu2020partpose} constructs a self-supervised training pipeline using the 3D-to-2D projection. PGP-human takes advantage of image pairs sampled from in-the-wild videos for training, which contain the same person performs different actions in different backgrounds. The model is trained to disentangle the appearance and pose information via mixing their features extracted from image pairs for image re-synthesis. In this process, the Puppet model~\cite{puppet_zuffi2015stitched} is adopted to transform the back-projected 2D pose into part segmentation maps for promoting appearance reconstruction. 
     
\subsubsection{Solving the Inherent Ambiguity Problem}:  
   Due to the inherent ambiguity in depth, a single 2D pose may correspond to multiple 3D poses, especially in the pipeline of lifting 2D pose to 3D. Therefore, various prior constraints are employed to determine the specific pose. 
   Many methods use temporal consistency and dynamics to solve the ambiguity of a single 2D pose. For example, the method RSTV~\cite{RSTV} directly regress from a spatio-temporal volume of bounding boxes to a 3D pose in the central frame. This approach achieves motion compensation by training two networks to  predict large body shifts between consecutive frames and then refine them. 
   Fang et al.~\cite{fang2018learning} explicitly incorporate the body prior (including kinematics, symmetry, and driven joint coordination) into the model via a hierarchy of bi-directional RNNs. In this way, 3D pose prediction is supervised to follow the body prior constraints and temporal dynamics. Besides, Lin et al. \cite{lin2017recurrent}, TP-Net \cite{rayat2018exploiting} and Lee et al.~\cite{lee2018propagating} develop sequence-to-sequence networks composed of LSTM units to estimate a sequence of 3D poses from 2D poses. The specific pose of each frame could be better determined via exploring the motion dynamics in a motion sequence.  Furthermore, to improve the computational efficiency, VideoPose3D~\cite{3dvideopose} and OANet~\cite{Cheng_2019_ICCV} adopt temporal convolution on 2D pose sequences to guarantee the temporal consistency. The full convolutional architecture enables efficient parallel computation. Differently, OANet uses a cylinder human model to generate the occlusion label, which helps the model to learn the collision between body parts. Moreover, Sharma et al.~\cite{sharma2019monocular} solve the ambiguity in a generative adversarial manner. They train a conditional VAE network to justify the rationality of the generated 3D-pose samples conditioned on the 2D  pose. In addition, ActiveMoCap~\cite{Kiciroglu_2020_CVPR} try to estimate the uncertainty of different predictions that are used to select the best output with lower ambiguity. It helps the model to learn the best viewpoints of 3D pose estimation. 
 
   Except the temporal consistency, some other methods attempt to solve the ambiguity via developing a representation closer to the body structure. For example, Xu et al.~\cite{DKA_Xu_2020_CVPR} and Li et al.~\cite{ETD_2020_CVPR} adopt a hierarchical bone representation, as introduced in Sec.~\ref{sec:representation}. The geometric dependence of adjacent joints is explicitly modeled in this hierarchical bone representation, which mainly focus on supervising the bone length and the joint direction. Benefited from the hierarchical bone representation, 3D body skeleton is separable and could be easily mixed up to synthesize new skeletons. Therefore, Li et al.~\cite{ETD_2020_CVPR} propose to enrich the pose-space of training data via mixing up different skeletons images. Training with expanded data helps to improve model generalization.
    
\subsubsection{Multi-person 3D Pose Estimation} 
In general cases, the real-world scenes always contain multiple persons. Similar to the challenges faced in multi-person 2D pose estimation, the 3D case also predicts the root depth and keypoint relative depth of each person. According to the processing pipeline, existing methods for multi-person 3D pose estimation can be roughly classified into two categories: 1) the top-down paradigm, and 2) the bottom-up paradigm.
The top-down methods first detect the person, then estimate the 3D pose of each one separately. While the bottom-up methods first detect the keypoints, then group them to form the 3D pose of each person.

	\textbf{{Top-down Methods}.}
  As a typical top-down paradigm, LCR-Net++~\cite{rogez2019lcr} is built on the common two-stage anchor-based detection framework. They first collect pose candidates from the anchor proposals and then determine the final output by score ranking. Similarly, the work in~\cite{moon2019camera} is also build on the anchor-based detection framework.
  They estimate the 3D absolute root localization and root-relative pose estimation with separate network branches from the detected person area and their bounding box location.
  
	\textbf{{Bottom-up Methods}.}
Following this paradigm, Zanfir et al.~\cite{zanfir2018deep} propose a bottom-up multi-stage framework for monocular multi-person 3D pose estimation. They first estimate the volumetric heatmaps from a single image to determine the 3D keypoint locations. Then the confidence scores of all possible connections between detected keypoints are predicted to form the limbs. Finally, they perform the skeleton grouping to assign the limbs to different persons. Moreover, to deal with the occlusion problem in multi-person scenes, Mehta et al.~\cite{mehta2018single} develop an Occlusion-Robust Pose-Maps (ORPM) to involve the redundant occlusion information in the part affinity maps. Besides, they propose the first multi-person 3D pose dataset, MuCo3DHP, which greatly promotes the development of this field. Furthermore, Fabbri et al.~\cite{fabbri2020compressed} propose to estimate the volumetric heatmaps in an encoder-decoder manner, and regress the multi-person 3D pose from them. The intermediate encoded features are supervised via the encoded feature of the ground truth keypoint heatmaps for compressing volumetric heatmaps. Differently, PandaNet~\cite{PandaNet_Benzine_2020_CVPR} is an anchor-based single-shot model for multi-person 3D pose estimation. It directly predicts the 2D/3D pose for each anchor position. In addition, SMAP~\cite{zhen2020smap} estimates multiple maps, representing the body root depth and part relative-depth at each position. To group the keypoints on heatmaps, they use the estimated depth to determine the association. HMOR~\cite{wang2020hmor} models multi-person interaction relations for better performance. It hierarchically estimates multi-person ordinal relations through instance-level, part-level, and joint-level. 
 
 \begin{figure}[!t]
	\centering
	\includegraphics[width=0.5\textwidth]{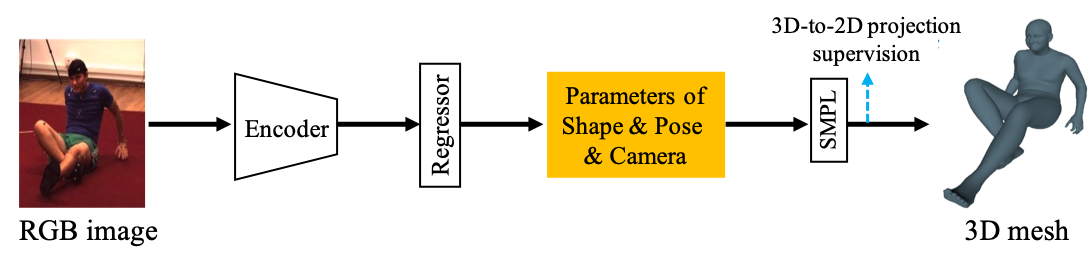}
	\caption{The representative framework of mesh-based 3D pose estimation via the SMPL model. }
	\label{fig:mesh_framework}
    \end{figure}

\subsection{Mesh-based 3D Pose Estimation}
 Mesh-based 3D pose estimation from a single image has attracted a lot of attention because it can provide extra body shape information beyond the locations of keypoints. Recent works in this community can be roughly classified into two categories. The first kind directly regresses the 3D human body shape from an input image, which represents the human body using a 3D mesh with thousands of vertices. For example, based on a pre-defined mesh topology, GraphCMR~\cite{Kolotouros_2019_CVPR} can regresses 3D mesh vertices using graph convolutional neural networks (GCNNs). Similarly, Pose2Mesh~\cite{Pose2Mesh} proposes a cascaded model using GCNNs. Differently, I2L-MeshNet~\cite{I2L-MeshNet}  proposes an image-to-lixel (line+pixel) prediction network, which predicts the per-lixel likelihood on 1D heatmaps to regress each mesh vertex coordinate.

Different from mesh vertex regression,  the other kind methods take the statistical 3D human model, like SMPL~\cite{smpl}, as the representation, which can bring the strong geometric prior. In this way, they formulate the 3D pose estimation as estimating the SMPL pose and shape parameters.  As shown in Fig.~\ref{fig:mesh_framework}, the general framework is to directly estimate the camera and SMPL parameters from a single-person 2D RGB image. As introduced in Sec.~\ref{sec:representation}, we can derive 3D human body mesh from SMPL parameters, and regress 3D keypoints from the mesh.  In this community, most works focus on how to 1) solve the 3D data shortage problem, 2) facilitating more proper representations for mesh-based 3D pose estimation, and 3) dealing with multi-person cases in real application scenes.


\subsubsection{Solving the Lack of Data}
    Due to the data shortage, we are supposed to use all 2D/3D pose datasets available for supervision. Many approaches develop various loss functions to supervise different aspects of the estimated body mesh. Human Mesh Recovery (HMR)~\cite{hmr} exploits the way of learning from unpaired data from 2D pose and 3D motion capture (MoCap) datasets. If only learning from 2D poses without depth-wise supervision, it will result in unreasonable 3D pose and shape. Instead, they use the MoCap data to supervise the rationality of the estimated parameters in a generative adversarial manner. A discriminator is developed to determine whether the estimated SMPL pose and shape parameters are reasonable. 
    
    To guide the model to explicitly learn from the existing data, some approaches supervise with the inherent properties of 2D images. To learn from the temporal dynamics, a 3D human dynamics model~\cite{kanazawa2018learning} is trained to estimate 3D poses of the current, past, and future frames. To transfer the static images to a motion sequence, a  hallucinator is learned to estimate the features of the past and future motion for synthesis. To learn from the temporal smoothness, Kocabas et al.~\cite{kocabas2019vibe} develop a temporal network named VIBE. Following HMR, they employ a motion discriminator to supervise the rationality of the predicted motion sequences in a generative adversarial manner. Specifically, via Gated Recurrent Units (GRUs), the SMPL parameters which describe the motion sequence are mapped to a latent representation at each time step. In addition to exploiting temporal information, TexturePose~\cite{pavlakos2019texturepose} utilizes the appearance consistency of the same person among multiple viewpoints or adjacent video frames for supervision.
    Body textures are mapped from 2D images to UV maps, which semantically align the multi-view or sequential textures. Only visible parts among multiple UV maps of each person are supervised with texture consistency.
    
    Moreover, there are some methods proposed to develop more detailed supervision. 
    For example, HoloPose~\cite{Guler_2019_CVPR} propose a multi-task network that  estimates DensePose~\cite{Guler2018DensePose}, 2D and 3D keypoints, along with the part-based 3D reconstruction. An iterative refinement method is proposed to improve the alignment between the model-based 3D estimates of 2D/3D keypoints and DensePose.  
    Besides, Human Mesh Deformation (HMD)~\cite{zhu2019detailed} utilizes additional information, including body keypoints, silhouettes, and per-pixel shading, to refine the estimated 3D mesh. Via a hierarchical mesh projection and deformation refinement, body mesh is well aligned with the person in the input 2D image. 
    SMPLify~\cite{keep} proposes to estimate 3D human mesh by fitting the SMPL model to the predicted 2D keypoints and minimizing the re-projection error. The SMPL oPtimization IN the loop method try to combine the advantages of regression-based and optimization-based methods~\cite{kolotouros2019spin}. They utilize the SMPLify to refine the estimated results in the training loop to provide additional 3D supervision. 
    
\subsubsection{Model Representations}
	Considering that regressing all parameters from a global image feature is ambiguous for the complex body mesh, more and more researchers focus on exploring more proper representations for mesh-based 3D pose estimation. For example, GraphCMR~\cite{Kolotouros_2019_CVPR} utilizes a graph-based representation for the 3D body mesh.
    Via a graph convolution network (GCN), the 3D location of each mesh vertice is estimated at each node. Then the SMPL parameters can be estimated from these vertices. Based on DensePose, DenseRaC~\cite{Xu_2019_ICCV} uses the estimated IUV map as the intermediate representation to estimate SMPL parameters. Specifically, a differential renderer is employed to render the estimated body mesh back to the IUV map and make a comparison with the input for supervision. Sun et al.~\cite{sun2019dsd-satn} develop a skeleton-disentangled representation using the bilinear transformation to tackle the feature coupling problem of 2D pose and the other details. They also employ a transformer-based network to learn the temporal smoothness, where an unsupervised adversarial training strategy is developed to learn the motion dynamics by ordering the shuffled frames.
   
    Different from recovering a single-body-part 3D mesh, some works extend the research into the representation for multi-body-part 3D mesh recovery.  
    For example, SMPL+H~\cite{smplh} integrates a 3D hand model into the SMPL body model to jointly recover the 3D mesh of the body and hands. Xiang et al.~\cite{monocualar} propose the method MTC to use separate CNN networks for estimating body, hands, and face, and then jointly fit the Adam~\cite{totalcapture} model to the outputs of all body parts. SMPL-X~\cite{smplx} combines FLAME head model~\cite{flame} with SMPL+H, and learns the pose-dependent blending shapes by fitting the model to 3D scans data. SMPLify-X~\cite{smplx} is proposed to recover the human whole body 3D mesh by iteratively fitting SMPL-X to 2D keypoints of face, hands, and body.  
   
\subsubsection{Multi-person 3D Mesh Recovery}
  Although great progress has been made on monocular 3D human pose and shape estimation for the single person scene, it is crucial to deal with multi-person cases with the truncation, environmental occlusion, and person-person occlusion. Existing multi-stage methods equip the single-person pipeline with a 2D person detector to handle multi-person scenes. Different from 2D/3D keypoints estimation that only estimates dozens of body joints, recent works also attempt to explore the particularity of 3D mesh recovery. 
    For example, Zanfir et al.~\cite{zanfir2018monocular} propose to use the natural scene constraints in multi-person scene. To get the initial 3D body meshes, they fit the SMPL model to the 3D poses and their semantic segmentation estimated from the image. To exclude the case of volume occupancy, they put a collision constraint into the objective function. Meanwhile, the ground plane is estimated to model the interactions between the plane and all human subjects. Furthermore, Jiang et al.~\cite{Jiang_2020_CVPR} propose to use the coherent reconstruction of multiple humans (CRMH) for multi-person 3D mesh recovery. They build their method based on Faster-RCNN~\cite{ren2015faster}, where the RoI-aligned features are used to predict the SMPL parameters. Specifically, they develop a differentiable interpolation loss to avoid collision between body meshes. Besides, to learn the correct depth ordering between multiple persons, they supervise the rendering of multi-person body meshes by instance segmentation. Recently, Sun et al.~\cite{CenterHMR} present a real-time Center-based Human Mesh Recovery network (CenterHMR) that is a novel bottom-up single-shot method. The model is trained to simultaneously predict two feature maps, which represent the location of each human body center and the corresponding parameter vector of 3D human mesh at each center, respectively. The explicit center-based representation guarantees the pixel-level feature encoding. The 3D mesh result of each person is estimated from the features centered at the visible body parts, which improves the robustness under occlusion. Besides, to deal with crowded cases with severe overlapping, an occlusion-aware center representation is proposed in this paper. 

 \textbf{ \textit{In summary}}, both the skeleton-based methods and mesh-based methods of 3D pose estimation have been greatly improved. With the aid of 2D pose estimation and the methods of un-supervision or weak-supervision, the problem of lack of 3D data can be solved to a certain extent. Meanwhile, with more exploration for the representations of 3D body, the mesh-based 3D pose estimation moves towards more accurate and efficient directions.
   

	\begin{figure*}[]
	\centering
	\captionsetup{justification=centering}
	\includegraphics[width=0.95\textwidth]{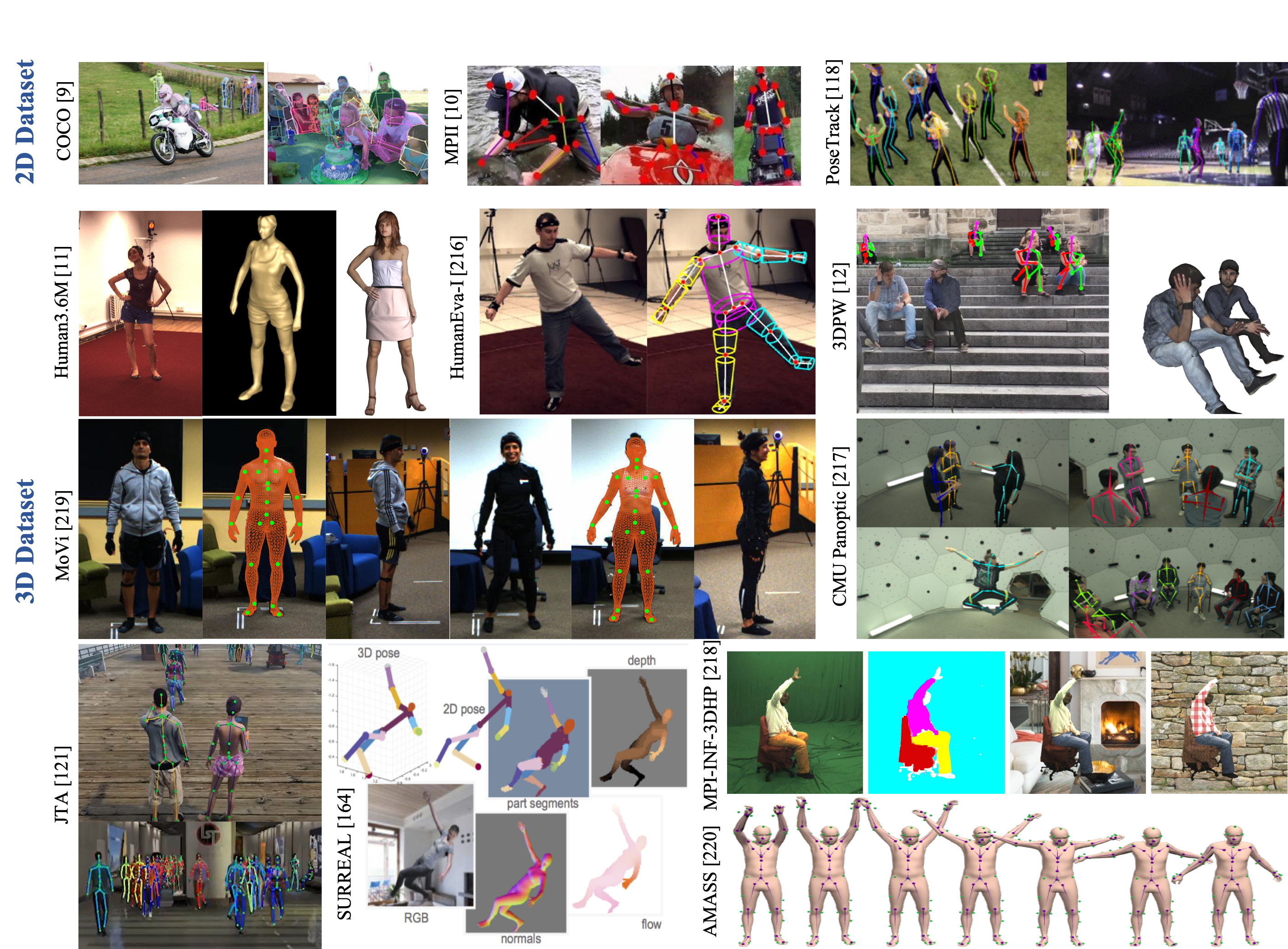}
	\caption{Some example annotated images selected from the monocular 2D (first row) and 3D pose benchmarks.}
	\label{fig:dataset_examples}
        \vspace{-4mm}
    \end{figure*}
	
	\section{Evaluation Metrics and Datasets}

	\subsection{Evaluation Metrics}
	
\subsubsection{Evaluation Metrics of 2D Pose Estimation}

The evaluation of 2D pose estimation aims to measure the accuracy of the predicted 2D locations. According to the characteristics of datasets, the widely used evaluation metrics include the Percentage of Correct Parts (PCP)~\cite{PCP}, Percentage of Correct Keypoints (PCK)~\cite{PCK}, and Average Precision (AP)~\cite{PCK, Dataset_COCO}, which are introduced as following.

\textbf{Percentage of Correct Parts (PCP)}~\cite{PCP} is proposed to measure the accuracy of body part prediction. The body part prediction is accurate if the estimated two endpoints of the corresponding limb are within a threshold ($50\%$) of the ground-truth endpoints. Specifically, PCPm in~\cite{Dataset_MPII} is defined by using $50\%$ of the mean ground-truth segment length over the entire test set as a matching threshold of PCP. However, PCP has a drawback that the foreshortening affects the correct measurement of body parts in different views and ranges. 

\textbf{Percentage of Correct Keypoints (PCK)}~\cite{PCK} is a widely used metric to measure the accuracy of 2D keypoints prediction. In~\cite{PCK}, the threshold for measuring keypoints to the ground-truth is defined as a fraction of the person bounding box size. Similarly, PDJ, the Percentage of Detected Joints, sets the threshold as the pixel radius that is normalized by the torso height of each test sample~\cite{Dataset_FLIC}.  PCKh@0.5 is a slight modification of the PCK. It adopts the matching threshold as $50\%$ of the head segment length of the testing person. By using the head size as a reference, PCKh makes the measurement articulation independent. By altering the threshold percentage, Area Under the Curve (AUC) can be generated to further evaluate the power of different pose estimation algorithms. 

\textbf{Average Precision (AP)}, first called the average precision of keypoints (APK), is proposed in~\cite{PCK} to measure pose estimation in a real system which has no annotated bounding boxes at test time. A detected keypoint candidate is considered to be correct (true positive) if it is within a threshold of the ground-truth. Each keypoint separately calculates its correspondence with the ground-truth poses. AP correctly penalizes both missed detections and false positives. In~\cite{Dataset_COCO}, for multi-person pose estimation,  AP is calculated by measuring the Object Keypoint Similarity ($\operatorname{OKS}$). Simialr to IoU in object detection, $\operatorname{OKS}$ measures the similarity between the predictions and the ground-truths:
\begin{equation}
\operatorname{OKS} = \frac{\sum_{i}\exp(-d_i^2/2s^2k_i^2)\delta(v_i > 0)}{\sum_i \delta(v_i > 0)},
 \end{equation}
where  $d_i$ is the Euclidean distance between the detected keypoint and the corresponding ground truth, $v_i$ is the ground-truth visibility flag, $s$ is the person scale, and $k_i$ is a per-keypoint constant that controls falloff. For each keypoint the $\operatorname{OKS}$ ranges between 0 and 1. 

Given the $\operatorname{OKS}$ over all labeled keypoints, the average precision (AP) and average recall (AR) can be computed. By tuning $\operatorname{OKS}$ values, the precision-recall curve can be calculated. $\operatorname{AP}$ and $\operatorname{AR}$ at different OKS can throughly reflect the performance of the testing algorithms. For COCO dataset~\cite{Dataset_COCO}, 10 metrics are used for evaluating the performance of a keypoint detector, including $\operatorname{AP}^{0.5}$ ($\operatorname{AP}$ at $\operatorname{OKS}  = 0.50$),  $\operatorname{AP}^{0.75}$, $\operatorname{AP}$ ( the mean of $\operatorname{AP}$ scores at 10 values, $\operatorname{OKS} = 0.50 : 0.05 : 0.95$), $\operatorname{AP}^{M}$ for medium objects, $\operatorname{AP}^{L}$ for large objects, $\operatorname{AR}^{0.5}$, $\operatorname{AR}^{0.75}$, $\operatorname{AR}$, $\operatorname{AR}^{M}$ for medium objects, $\operatorname{AR}^{L}$ for large objects.
The metrics of $\operatorname{AP}$ and $\operatorname{AR}$ are useful for understanding which keypoints are more difficult than others. They have been widely used as the evaluation metric for multi-person pose estimation.

	\subsubsection{Evaluation Metrics of 3D Pose Estimation}
	
	\textbf{The Mean Per Joint Position Error (MPJPE)} is the most widely used evaluation metrics of 3D pose estimation. It measures the average Euclidean distance from the 3D pose predictions to the ground truth in millimeters. The predictions and ground truth keypoints are aligned by the pelvis keypoint for comparison. 
	
	\textbf{Procrustes Aligned MPJPE (PA-MPJPE)} is a modification of MPJPE, which can be obtained by rigidly aligning the predicted pose with ground truth in millimeters. It is also called as the reconstruction error. Through Procrustes alignment, the effects of translation, rotation, and scale are eliminated, which makes PA-MPJPE focus on evaluating the accuracy of the reconstructed 3D skeleton. 
    
   \textbf{3D PCK \& AUC.}  3D PCK is the 3D version of the PCK metric. The threshold of success prediction is usually set to 50mm or 150mm in different methods. Correspondingly, the AUC, which is the total area under the PCK-threshold curve, is calculated by computing PCKs by varying the threshold from 0 to 200 mm.
   
   \textbf{The Mean Per Joint Angle Error (MPJAE)} measures the angle between the predicted keypoint orientation and the ground truth orientation in degrees. The orientation difference is measured as the geodesic distance in SO(3). The detailed definition can be founded in~\cite{pons2014human}.  Besides, only the angles of four limbs and the root are used for evaluation.

\textbf{Procrustes Aligned MPJAE (PA-MPJAE)} measures the MPJAE normalized by the rotation matrix on all predicted orientations. The rotation matrix is obtained from the Procrustes alignment. Similarly, PA-MPJAE neglects the global mismatch.
    \vspace{-4mm}
	
	\subsection{Datasets}

The rapid development of the related datasets boost the development of deep learning-based pose estimation methods. Public available datasets provide training sources and fair comparison for different methods. Considering the dataset scale and diversity of poses and scenes, in this section we introduce the representative datasets in recent years.  Most of them are high-quality and large-scale datasets with good annotations in different shooting scenes. 

\subsubsection{2D Pose Datasets}
we introduce 2D datasets according to the categories of 1) image-level or video-level, and 2) single person or multiple persons. The dataset summary is given in Table~\ref{table:dataset_2d}. Part of the example images is shown in Fig.~\ref{fig:dataset_examples}.

\begin{table*}[t]
		\caption{2D pose estimation datasets.  Kpts is short for keypoints. For the train/val/test set, the number of images in image-based datasets and the number of videos in video-based datasets are listed. }
		  \setlength\tabcolsep{4pt}
	\centering
			\label{table:dataset_2d}
			\footnotesize
            \begin{tabular}{c|c|c|c|c|c|c|c|c}
				\hline
                                  Dataset & Type & Year & Image/Video & Num of Kpts  & Train set & Val set & Test set & Evaluation  \\
				\hline
                                  LSP~\cite{Dataset_LSP}                          & Single &  2010 & Image  &  14  &  1K  &  0 &  1K &  PCK \\
                                  FLIC~\cite{Dataset_FLIC}                       &   Single & 2013 & Image & 10  & ~5K  & 0& ~1K & PCK \\
                                  MPII Single Person~\cite{Dataset_MPII} & Single & 2014 & Image& 16  & 29K & 0 & 12K & PCK \\
                                  MPII Multi-Person~\cite{Dataset_MPII} & Multiple & 2014 & Image& 16  & 3.8K & 0 & ~1.7K & PCK \\
                                  COCO 17~\cite{Dataset_COCO}               & Multiple & 2016 & Image& 17 & 57K & 5K & 20K & mAP \\
                                  AI-Challenger~\cite{Dataset_AIC}        & Multiple & 2017 & Image & 14 &       210K &  30K &  60K  & mAP \\
                                  CrowdPose~\cite{19MP_li2019crowdpose} &  Multiple & 2019 & Image & 14 &  10K & 2K & 8K & mAP \\
                                  HiEve~\cite{Dataset_HIE}                   &   Multiple & 2020 & Both &  14&   33K & 0 & 16K & mAP \\
                                  \hline
                                  J-HMDB~\cite{Dataset_JHMDB}        & Single & 2013  & Video & 15   & 0.6K & 0 & 0.3K & PCK \\
                                  Penn Action~\cite{Dataset_PennAction} & Single & 2013  & Video & 13 & ~1K & 0& ~1K  & PCK \\
                                  PoseTrack~\cite{VOP18_Posetrack18} & Multiple & 2017 &  Video & 15  &    0.29k & 0.05k & 0.2K & mAP\\ 
				\hline
			\end{tabular}
	\end{table*}

 \textbf{$\bullet$} { \textbf{Image-level 2D Single Person Dataset:}}

\textbf{Leeds Sports Pose (LSP) Dataset}~\cite{Dataset_LSP} is collected from Flickr using the tags of eight sport activities (athletics, badminton, baseball, gymnastics, parkour, soccer, tennis, and volleyball). The dataset contains 2,000 images, of which 1,000 images for training and the rest 1,000 images for testing. Each person is labeled by 14 keypoints of the full body. Compared with those newly released datasets, LSP is relatively small-scale. It is a initial performance evaluation for single person pose estimation methods. 

\textbf{Frames Labeled in Cinema (FLIC) Dataset}~\cite{Dataset_FLIC}  contains 5,003 images collected from Hollywood movies. They run the person detector Poselets~\cite{bourdev2009poselets} on every tenth frame of 30 movies. Originally 20K candidates are selected be labeled by the crowdsourcing marketplace Amazon Mechanical Turk with 10 upper-body keypoints. Images with persons occluded or severely non-frontal are filtered out. Finally, 1,016 images are selected as the testing set.

	\begin{table*}[t]
		\caption{Performance of the representative 2D single pose estimation methods on the MPII test set.}
	\setlength\tabcolsep{4pt}
	\centering
			\label{table:mpii}
			\footnotesize
            \begin{tabular}{c|c|c|c|l|c}
		\hline
		\textbf{Method }&\textbf{Publication} & \textbf{Backbone}&\textbf{ Input size }& \textbf{Keywords of Network }& \textbf{PCKh@0.5} \\
		\hline
      Tompson et al.~\cite{14SP_TompsonJLB14} &  NeurIPS'14     &  AlexNet & $320\times240$    & Keypoint heatmap, multi-resolution, MRF spatial model &   79.6 \\ 
      Carreira et al.~\cite{16SP_CarreiraAFM16} & CVPR'16 &  GoogleNet &   $224\times224$   & Self-correcting model   with   iterative update  &       81.3\\
      Tompson et al.~\cite{SP15_tompson2015efficient} &  CVPR'15   &   AlexNet & $256\times256$    & Position refinement model to predict joint offset location   &      82.0         \\
      Hu  et al.~\cite{SP15_hu2016bottom}    & CVPR'16   & VGG & $224\times224$    &   Hierarchical Rectified Gaussian model &      82.4\\
      Pishchulin et al.~\cite{16BU_pishchulin2016deepcut} & CVPR'16  &  VGG  &  $340\times340$    & Combine detection and pose estimation based on Fast R-CNN &   82.4 \\
      Lifshitz et al.~\cite{16SP_LifshitzFU16} &  ECCV'16   & VGG &    $504\times504$ & Each pixel in the image votes for the positions of keypoints  &   85.0 \\
      Gkioxary et al.~\cite{16SP_GkioxariTJ16} & ECCV'16 & InceptionNet   & $299\times299$ & Chained  model, each keypoint relies on its previous ones    & 86.1\\
      Sun et al.~\cite{sun2017compositional}  & ICCV'17 & ResNet-50& $224\times224$ & Add bone-based representation as constraints  & 86.4 \\ 
      Insafutdinov et al.~\cite{16BU_Insafutdinovdeepercut} & ECCV'16 &  ResNet-152 &$340\times340$&     Keypoint geometric and appearance constraints & 88.5 \\
      Wei et al.~\cite{SP16-CPM} &CVPR'16    & CPM & $368\times368$   &    Convolutional Pose Machines,  intermediate supervision   & 88.5 \\
      Newell et al.~\cite{16SP_NewellYD16} & ECCV'16 &Hourglass & $256\times256$  &  Hourglass model, intermediate supervision    & 90.9 \\
      Sun et al.~\cite{SP17_sun2017human} & ICCV'17 &Hourglass &$340\times340$ & Two-stage normalization,multi-scale supervision and  fusion &91.0 \\
      Tang et al.~\cite{18SP_TangPGWZM18} & ECCV'18 & Hourglass   &$256\times256$ &  Order-K dense connectivity, quantification  to low bit  &91.2 \\
      Luvizon et al.~\cite{19SP_LuvizonTP19} & CG'19 &  Hourglass & $256\times256$ & Multi-stage, soft-argmax on heatmaps & 91.2 \\
      Nie  et al.~\cite{19_ContextualRefine} & TIP'19 &  Hourglass &$384\times384$&   Hierarchical contextual refinement network  &  $91.2$ \\    
      Chu et al.~\cite{17SP_ChuYOMYW17} & CVPR'17 &   Hourglass &  $256\times256$ &  multi-resolution attention maps, Hourglass Residual Units   &  91.5\\
Chen et al.~\cite{17SP_ChenSWLY17} & ICCV'17  & En/Decoder    &  $256\times256$ & GAN, multi-task for poses and occlusion parts, structure-aware  & 91.9 \\
Yang et al.~\cite{17SP_YangLOLW17} &  ICCV'17 & Hourglass &  $256\times256$ &   pyramid residual module to learn various scales       & 92.0\\
Ke et al.~\cite{18SP_KeCQL18}  & ECCV'18 &Hourglass &  $256\times256$ &  Multi-scale supervision, structure-aware loss, keypoint masking& 92.1 \\
Tang et al.~\cite{18SP_TangYW18}    &  ECCV'18 &Hourglass &  $256\times256$ &  Hierarchical compositional model, bone-based representation     & 92.3 \\
Sun et al.~\cite{19MP_HrnetXLW19} & CVPR'19 & HRNet &$256\times256$ &  Maintain high-resolution maps, multi-branch/scale fusion     &92.3 \\
Zhang et al.~\cite{19SP_abs-1901-01760} & ArXiv'19 & Hourglass   &$256\times256$ &   Cascade fusion, graph neural network for refinement &   92.5 \\
 Tang et al.~\cite{19SP_TangW19}  & CVPR'19 & Hourglass &  $256\times256$ &   data-driven keypoint grouping, part-based branching network & 92.7 \\
   				\hline
			\end{tabular}
			\vspace{-5mm}
	\end{table*}   
\textbf{MPII Dataset}~\cite{Dataset_MPII} is a large-scale dataset containing rich activities and diversity capture environments, both indoor and outdoor. It is collected from 3,913 videos spanning 491 different activities from YouTube. A total of 24,920  frames are extracted from the collected videos. The annotations are conducted by in-house workers on Amazon Mechanical Turk (AMT). The annotations include 2D locations of 16 keypoints, full 3D torso and head orientation, occlusion labels for keypoints, and activity labels. Adjacent video frames are also available for motion information.  Finally, 40,522 people are labeled, of which 28,821 people are used for training and 11,701 for testing. MPII dataset has been widely used for pose estimation and other pose related tasks. 
Table.~\ref{table:mpii} shows the state-of-the-art methods evaluated on the MPII test set. Since the posture is relatively easy, the accuracy of detected 2D keypoints is high and the performance is close to saturation.

	\begin{table*}[t]
	\caption{Performance of the representative 2D multi-person pose estimation methods on the COCO test-dev set. For bottom-up methods, the reported results use multi-scale testing.}
	\setlength\tabcolsep{4pt}
	\centering
			\label{table:coco_test_dev}
			\footnotesize
            \begin{tabular}{c|c|c|c|c|c}
				\hline
				\textbf{Method} &\textbf{Publication} &\textbf{ Backbone}& \textbf{ Input size }& \textbf{Keywords of Network} &{\textbf{mAP}}  \\
				\hline
				\multicolumn{6}{c}{\textbf{Bottom-up: keypoint detection and grouping}}\\
				\hline
				OpenPose~\cite{openpose} &CVPR'17 & CMU-Net & $368\times368$ & Multi-stage, part affinity fields. &$61.8$ \\		
				Asso. Emb.~\cite{17BU_newell2017associative} &NeurIPS'17 & Hourglass & $512\times 512$ &  Hourglass for pose, associative embedding for grouping. & $65.5$ \\
				PersonLab~\cite{18BU_papandreou2018personlab} & ECCV'18& ResNet & $801\times 801$ &Multi-task, short/mid/long-rang offsets for grouping.
				&$68.7$\\
				MultiPoseNet~\cite{18BU_kocabas2018multiposenet} & ECCV'18 &  ResNet &$480\times480$&   Multi-task, Pose Residual Network. &$69.6$\\
				PifPaf~\cite{19BU_kreiss2019pifpaf} &CVPR'19& ResNet& $401\times401$ & Part Intensity Field and Part Association Field& $66.7$ \\
				Li et al.~\cite{2020BU_li2020simple} & AAAI'20 & IMHN & $512\times 512$  & Hourglass network with spatial/channel attention & $68.1$  \\
				HigherHRNet~\cite{cheng2020higherhrnet} &CVPR'20& HRNet  & $640\times 640$ & Feature pyramid, associative embedding. & $70.5$   \\				
				\hline 
				\multicolumn{6}{c}{\textbf{Top-down: human detection and single-person keypoint detection}}\\
				\hline
				Mask-RCNN~\cite{17BU_he2017mask} & ICCV'17 & ResNet& $800\times800$ & Faster-RCNN, multi-task, RoIAlign. & $63.1$   \\
				G-RMI~\cite{17MP_papandreou2017towards} &CVPR'17 &  ResNe & $353\times257$ &  Full convolutions, keypoint offsets, keypoint  NMS. &$64.9$ \\
				Integral Regre.~\cite{sun2018integral} &ECCV'18 & ResNet & $256\times256$ &  Intergral loss and keypiont regression.
				&$67.8$ \\
				CPN~\cite{18MP_chen2018cascaded} &CVPR'18 & ResNet& $384\times288$ & GlobalNet and RefineNet, online hard keypoint mining. 
				& $72.1$ \\
				RMPE~\cite{17MP_fang2017rmpe} &ICCV'17 & PyraNet & $320\times256$ & Spatial transformer, parametric NMS,  proposals generator.
				&$72.3$\\
				CFN~\cite{17SP_HuangGT17} &ICCV'17 & Inception & $448\times448$ & Multi-level supervision and fusion, coarse to fine.
				& $72.6$\\
				SimpleBaseline~\cite{18MP_xiao2018simple} &ECCV'18 & ResNet&$384\times288$  & Deconvolution pose head network.
				&${73.7}$ \\
                CSM+SCASRB~\cite{19MP_su2019multi} & CVPR'19 &ResNet & $384\times288$ &   Channel shuffle module, spatial channel-wise attention.  & $74.3$ \\
				HRNet-W$48$~\cite{19MP_HrnetXLW19} & CVPR'19 & HRNet& $384\times 288$ & Maintain high-resolution, multi-branch/scale fusion.
				& ${75.5}$ \\
                MSPN~\cite{18MP_li2019rethinking} & arXiv'19 & MSPN & $384\times 288$ & Multi-stage feature aggregation,  coarse-to-fine loss. & $76.1$ \\
                DARK~\cite{20_DARK} & CVPR'20 & HRNet & $384\times288$ & Distribution-aware coordinate representation of keypoint. & $76.2$ \\
                UDP~\cite{20_UDP} &  CVPR'20 & HRNet & $384\times288$ & Unbiased data processing. & $76.5$ \\
                PoseFix~\cite{19MP_PoseFix} &  CVPR'19 & HR+ResNet & $384\times288$ & Pose refinement,
error statistics to generate synthetic poses. & $76.7$ \\
                Graph-PCNN~\cite{20_MP_Graph-PCNN} & CVPR'20 & HRNet & $384\times288$ & Two-stage
graph-based and model-agnostic framework.
& $76.8$ \\
                RSN~\cite{20_MP_DelicateLearning} &CVPR'20 & 4-RSN & $384\times288$ & Residual steps
network, pose
refine machine.
 & $78.6$  \\    
				\hline
			\end{tabular}
	\end{table*}
	
\textbf{$\bullet$} \textbf{Image-level 2D Multi-Person Dataset}:

\textbf{Microsoft Common Objects in COntext (MSCOCO)  Dataset}~\cite{Dataset_COCO} contains annotations for object detection,  panoptic segmentation, and keypoint detection. The images are collected from websites including Google, Bing, and Flickr. The annotations are performed by workers on Amazon's Mechanical Turk (AMT). The dataset contains over 200K images and 250K person instances. Along with the dataset, the Challenge of COCO Keypoint Detection is held every year since 2016. The dataset has two versions. The difference is the split of training and validation set. In the latest 2017 version, the training/val images split is 118K/5K instead of the previous 83K/41K. The test set contains 20K images and the annotations are hold out by the official testing server. Besides, 120K unlabeled images are also released that follow the same class distribution as the labeled images. They may be used for semi-supervised learning. For keypoint detection, 17 keypoints are labeled along with the visibility tag, bounding box, and body segmentation area.  COCO dataset has been a widely used evaluation benchmark, and served as auxiliary data for pose related tasks such as action recognition and person ReID.
Table.~\ref{table:coco_test_dev} shows the performance of the state-of-the-art methods on the COCO test set. RSN~\cite{20_MP_DelicateLearning} achieves $78.6$ mAP showing the superiority of the top-down methods. With the improvement of the network backbone and the keypoint grouping method, the bottom-up methods have rapidly developed.  HigherHRNet~\cite{cheng2020higherhrnet} obtains $70.5$ mAP. The bottom-up methods likely have the potential to achieve comparable performance with the top-down ones.  

\textbf{AI-Challenger Dataset}~\cite{Dataset_AIC}, which is also referred as the Human skeletal system Keypoint Detection Dataset (HKD), contains 300K high-resolution images
for keypoint detection and Chinese captioning, and 81,658 images for zero-shot recognition. 
The large-scale dataset has multiple persons and various poses. Each person is labeled with a bounding box and 14 keypoints.
The whole dataset is divided into the training set, validation set, test A set, and test B set with 210K, 30K, 30K, and 30K images, respectively.
Due to its large scale, high resolution, and rich scenes,  AI-Challenger Dataset has been widely used as an auxiliary dataset for 2D/3D pose estimation network training and pose related tasks.

\textbf{CrowdPose Dataset}~\cite{19MP_li2019crowdpose} is intended for better evaluating human pose estimation methods in crowded scenes. The images are collected from
MSCOCO (person subset), MPII, and AI Challenger by measuring the Crowd Index. Crowd Index is defined to evaluate the crowding level of an image. 30K images are analyzed via the Crowd Index and finally 20K high-quality images are selected. Next, 14 keypoints and full-body bounding boxes are annotated for about 80K persons. The training, validation, and testing subset are split in proportion to 5:1:4. Since the detection for either person bounding boxes or keypoints in crowd scenes is relatively hard, the CrowdPose dataset is still challenging in the multi-person pose estimation community. 

\textbf{$\bullet$} \textbf{Video-level 2D Single Person Dataset}:

\textbf{J-HMDB Dataset}~\cite{Dataset_JHMDB}, short for joint-annotated HMDB, is a subset of the HMDB51 database~\cite{HMDB} that contains over 5,100 clips of 51 human actions. J-HMDB Dataset contains 928 clips with 21 action categories. Each action class contains 36-55 clips. Each clip includes 15-40 frames. 31,838 images are annotated via a 2D puppet model~\cite{puppetflow} on Amazon Mechanical Turk. Up to 15 visible body keypoints are labeled, along with the scale, viewpoint, segmentation, puppet mask, and puppet flow. The ratio for the number of training and testing images is roughly 7:3. J-HMDB dataset has been widely used in the task of pose estimation in videos and action recognition.

\textbf{Penn Action Dataset}~\cite{Dataset_PennAction} is another unconstrained video dataset that contains 2,326 video clips covering 15 actions. 
The training set and testing set both have 1,163 video clips. 
The dataset contains various intra-class actor appearances, action execution rate, viewpoint, spatio-temporal resolution, and complicated natural backdrops. 
Annotation is conducted via a semi-automated video annotation tool deployed on Amazon Mechanical Turk. 
Each person is annotated with 13 keypoints with 2D coordinates, visibility, and camera viewpoints.

\begin{table*}
  \setlength\tabcolsep{4pt}
    \caption{Performance of the representative multi-person pose estimation and  tracking methods on PoseTrack 2017 test set. }
  \centering
  	\footnotesize
  \label{tab:final-MOTA_mAP-17-test}
      \begin{tabular}{c | c |c |c | c}
      \hline
     \multirow{2}{*}{\bf{Method} }& \multirow{2}{*}{\bf{Publication} }& \multirow{2}{*}{\bf Keywords} &  \bf Total   &  \bf{Total }  \\
     & & & \textbf{mAP}&\textbf{ MOTA} \\
     \hline
     \multicolumn{5}{c}{\textbf{Bottom-up: keypoint detection and grouping}}\\
     \hline
     ArtTrack \cite{VOP17_InsafutdinovAPT17} &CVPR'17 & Spatio-temporal clustering and grouping for keypoints & 59.4 & 48.1 \\  
    PoseTrack \cite{VOP17_Posetrack17} & CVPR'17 & Bottom-up pose estimation and spatial-temporal graph &  59.4  &48.4 \\
     JointFlow\cite{VOP18_jointflow}  & BMVC'18 &  Temporal flow fields to propagate  keypionts across frames &  63.4 &53.1 \\  
     TML++ \cite{VOP19_HwangLPK19} &  IJCNN'19 & Temporal flow maps for keypoints,  multi-stride frames sampling & 68.8 &  54.5\\
      STAF \cite{VOP19_RaajIHS19} &  CVPR'19 & Spatial-temporal affinity fields across a sequence & 70.3 & 53.8 \\ 

     \hline
       \multicolumn{5}{c}{\textbf{Top-down: person detection, keypoint detection, and data association}}  \\
       \hline 
    Detect-Track~\cite{VOP18_Detect-and-Track18} & CVPR'18 & 3D Mask R-CNN for each clip, bounding box IOU for similarity metric &59.6  &51.8 \\
     PoseFlow~\cite{VOP18_PoseFlow} & BMVC'18 & Pose flow building in short clips,  pose flow NMS post-processing & 63.0& 51.0\\
           LightTrack~\cite{LightTrack} & CVPRW'20 & Combining single-person pose tracking 
and object tracking, Siamese GCN  &66.7  &58.0 \\
                 POINet~\cite{POINet} & ACM MM'19 & Unifying feature extraction and data association by ovonic insight network & 72.5  & 58.4 \\
      PGPT~\cite{PGPT} & TMM'20 & Combining detection and single object tracking, GCN for data association & 72.6  & 60.2\\
      KeyTrack~\cite{KeyTrack} & CVPR'20 & Transformer-based pose tracking, parameter-free refinement & 74.0 & 61.2 \\
      DetTrack~\cite{VOP20_wang2020combining} & CVPR'20 & 3D HRNet for each clip,  spatio-temporal merging & 74.1  & 64.1 \\
    Self-Sup.~\cite{20_VOP_Self-supervised} & ECCV'20 & Self-supervised keypoint correspondences & $74.2$& $60.0$ \\
      FlowTrack~\cite{18MP_xiao2018simple} & ECCV'18 & Bounding boxes propagation and optical flow-based temporal similarity  & 74.6  & 57.8 \\ 
   \hline
    \end{tabular}
\end{table*}
\textbf{$\bullet$} \textbf{Video-level 2D Multi-Person Dataset}:

\textbf{PoseTrack Dataset}~\cite{VOP17_Posetrack17,VOP18_Posetrack18} is the first large-scale multi-person pose estimation and tracking dataset. It is collected from the unlabelled videos in MPII Multi-Person Pose dataset~\cite{Dataset_MPII}. It has two versions, i.e., PoseTrack 2017 and PoseTrack 2018.  
PoseTrack 2017 contains 550 videos split into 292, 50, and 208 videos for training, validation, and testing, respectively.  Totally 23,000 frames are annotated with 153,615 pose labels. PoseTrack 2018 is its extended version. It contains 593 training videos, 170 validation videos, and 375 testing videos. For each video in the training set, the middle 30 frames are annotated. For the validation set and testing set, the middle 30 frames along with every four frames are annotated. 
The labels contain 15 2D keypoints, an unique person ID, and the head bounding box for each person. PoseTrack is challenging since the videos contain various pose appearance and scale variation, along with body part occlusion and truncation. It has been a widely used benchmark to evaluate multi-person pose estimation and tracking algorithms.

Table.~\ref{tab:final-MOTA_mAP-17-test} presents the performance of the representative methods on the PoseTrack 2017 test set. The multi-step top-down methods show superior performance over the bottom-up methods, while the latter ones are more efficient.
The pose estimation task only relies on the accuracy of keypoint prediction, while the pose tracking also needs a solid and robust data association scheme.
With the development of pose estimation and data association, pose tracking has the potential to achieve better performance with higher efficiency.

\textbf{Human-in-Events (HiEve) Dataset}~\cite{Dataset_HIE} is a large-scale video-based dataset for realistic events, especially for crowd and complex events. It contains 2D poses, actions, trajectory tracking, and pose tracking.
The dataset is collected from 9 realistic scenes containing 49,820 frames with  annotations for 1,302,481 bounding boxes, 2,687 track IDs, 56,643 actions (14 action categories), and 1,099,357 human 2D poses. The label for 2D pose contains 14 keypoints and filters out the heavy occlusion and small bounding box (less than 500 pixels). HiEve dataset is the largest scale human-centric dataset to date, which will be useful in many tasks for human behavior analysis. 

	\begin{table*}
         \setlength\tabcolsep{2pt}
		\caption{The details of the widely used 3D pose estimation datasets.}
		\label{3D pose datasets}
		\centering
         	\footnotesize
		\begin{tabular}{c | c| c | c c c c c | c}
			\hline
			\multirow{2}[1]{*}{\textbf{Datasets}} & \multicolumn{1}{c|}{\textbf{Total}}& \multicolumn{1}{c|}{\textbf{Camera}} & \multicolumn{5}{c|}{\textbf{Annotation Type}} & \multirow{2}[1]{*}{\textbf{Code Link}}\\
			\cline{4-8}
			& \textbf{Frames} & \textbf{Viewpoints} & \textbf{3DP} & \textbf{2DP} & \textbf{Mesh} & \textbf{ITW} & \textbf{Real} \\
			\hline
			Human3.6M\cite{h36m} & 3.6 M & 4 & $\checkmark$ & $\checkmark$ & $\checkmark$ &  & $\checkmark$ & http://vision.imar.ro/human3.6m/description.php \\
			HumanEva-I\cite{humaneva} & 0.037 M & 7 & $\checkmark$ & $\checkmark$ & & &  $\checkmark$ & http://humaneva.is.tue.mpg.de/ \\
            CMU Panoptic\cite{Joo_2015_ICCV} & 1.5 M & 31 & $\checkmark$ &  $\checkmark$ &  & & $\checkmark$ & https://virtualhumans.mpi-inf.mpg.de/3DPW/ \\
			3DPW\cite{3dpw} & 0.051 M & 1 & $\checkmark$ &  $\checkmark$ & $\checkmark$ & $\checkmark$ & $\checkmark$ & http://domedb.perception.cs.cmu.edu/ \\
			MPI-INF-3DHP\cite{mono-3dhp2017} & 1.3 M & 14 & $\checkmark$ & $\checkmark$ &   &  &  $\checkmark$ & http://gvv.mpi-inf.mpg.de/3dhp-dataset/ \\
			JTA\cite{JTA} & 0.46 M & 1 & $\checkmark$ & $\checkmark$ &   & $\checkmark$ &  & https://github.com/fabbrimatteo/JTA-Dataset \\
			Varol et al.\cite{surreal} & 6.0 M & 1 & $\checkmark$ & $\checkmark$ & $\checkmark$ &  &  & https://www.di.ens.fr/willow/research/surreal/data/ \\
			\hline
		\end{tabular}
            \vspace{-4mm}
	\end{table*}
	\subsubsection{3D Pose Datasets}
We introduce the widely used 3D single person datasets, multi-person datasets, and the analysis of benchmarks in this part. The widely used 3D pose benchmarks are summarized in Table~\ref{3D pose datasets} and the example images are shown in Fig.~\ref{fig:dataset_examples}.

  \textbf{$\bullet$} \textbf{3D Single Person Dataset}:

	\textbf{Human3.6M}~\cite{h36m} is the most widely used multi-view single-person 3D human pose benchmark. The dataset is captured in a 4m$\times$3m indoor space using 4 RGB camera, 1 time-of-flight sensor, and 10 motion cameras. It contains 3.6 million 3D human poses and the corresponding videos (50 FPS) in 15 scenarios, such as discussion, sitting on a chair, taking a photo, etc. Especially, both 3D positions and angles of keypoints are available. Currently, due to privacy concerns, only 7 subjects' data is available. For evaluation, videos are usually down-sampled by every 5-th/64-th frame for removing the redundancy. Methods are often evaluated on two common protocols for comparison. The first protocol is to train on 5 subjects (S1, S5, S6, S7, S8) and test on subject S9 and S11. The second protocol shares the same train/test set, but only evaluates the images captured in the frontal view. 
	
	\textbf{HumanEva-I}~\cite{humaneva} is a single-person 3D pose dataset captured from 3 camera views at 60 Hz. It contains 4 subjects performing 6 actions. Related methods are usually evaluated on 3 actions, walk, jogging, and boxing, performed by 3 subjects, S1, S2, and S3.
 
	\textbf{MPI-INF-3DHP}~\cite{mono-3dhp2017} is captured in a 14 camera studio using commercial marker-less motion capture device for acquiring the ground truth 3D pose. It contains 8 actors performing 8 activities. The RGB videos are recorded from a wide range of viewpoints. Over 1.3M frames are captured from all 14 cameras. Except for the indoor videos of a single person, they also provide MATLAB code to generate a multi-person dataset, MuCo-3DHP, via mixing up segmented foreground human appearance. 
With the provided body part segmentation, researchers can also exchange the clothes and the backgrounds using extra texture data.

    \textbf{MoVi~\cite{ghorbani2020movi}} is a large-scale single-person video dataset with 3D MoCap annotations. Different from Human3.6M and MPI-INF-3DHP, it contains more subjects (60 female  and 30 male). Each person performs a collection of 20 predefined actions and one self-chosen movement. 
The video synchronized with motion capture was taken from two perspectives, front and side. Except for the 3D pose annotations and camera parameters, MoVi also provides the SMPL parameters obtained via the MoSh++~\cite{AMASS:2019}.

\begin{figure*}[!t]
	\centering
	\includegraphics[width=0.9\textwidth]{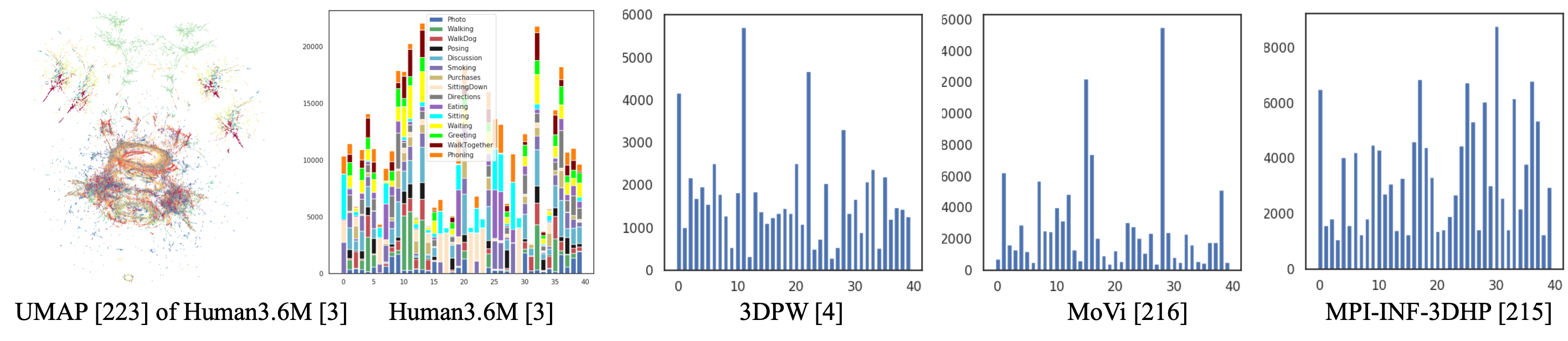}
	\caption{Pose space analysis for four 3D pose benchmarks: Human3.6M, 3DPW, MoVi, and MPI-INF-3DHP.}
	\label{fig:pose_space}
    \end{figure*}

	\begin{table*}
         \setlength\tabcolsep{4pt}
		\caption{Comparisons of 3D pose estimation methods on Human3.6M and HumanEva datasets.}
		\label{Comp_3dpose}
		\centering
         	\footnotesize
		\begin{tabular}{c | c | c | c | c | l }
			\hline
			\multirow{2}[1]{*}{\textbf{Method}} & \multirow{2}[1]{*}{\textbf{Publication}}   & \multicolumn{2}{c|}{\textbf{Human3.6M}}&\multicolumn{1}{c|}{\textbf{HumanEva-I}}& \multirow{2}[1]{*}{\textbf{Code Link}}\\
			\cline{3-4}
			
			& & \textbf{MPJPE }& \textbf{PMPJPE} & \textbf{PMPJPE} \\
			\hline
			Zhou et al.~\cite{sparseness} & CVPR'16 & 113.0 & - & - & https://github.com/chuxiaoselena/SparsenessMeetsDeepness\\
			Tome et al.~\cite{tome2017lifting} & CVPR'17 & 88.4 & - & - & https://github.com/DenisTome/Lifting-from-the-Deep-release\\
			C2F~\cite{pavlakos2017coarse} & CVPR'17 & 71.9 & 51.9 & 25.5 & https://github.com/geopavlakos/c2f-vol-demo \\
			Lin et al.~\cite{lin2017recurrent} & CVPR'17 & 73.1 & - & 30.9 & https://github.com/MudeLin/RPSM\\
			Martinez et al.~\cite{martinez2017simple} & ICCV'17 & 62.9 & 47.7  & 24.6 & https://github.com/una-dinosauria/3d-pose-baseline \\
			CHP~\cite{sun2017compositional} & ICCV'17 & 92.4 & 59.1 & - & - \\
			Fang et al.~\cite{fang2018learning} & AAAI'18 & 60.4 & 45.7 & 22.9 & - \\
			Pavlakos et al.~\cite{pavlakos2018ordinal} & CVPR'18 & 56.2 & 41.8 & 18.3 & https://github.com/geopavlakos/ordinal-pose3d \\
			 Luvizon et al.~\cite{18SP_LuvizonPT18} & CVPR'18 & 53.2 & - & - &  https://github.com/dluvizon/deephar\\
			Rhodin et al.~\cite{Rhodin_2018_CVPR} & CVPR'18 & 66.8 & 51.6 & - & - \\
			IHP~\cite{sun2018integral}  & ECCV'18 & 64.1 & 49.6 & - & https://github.com/JimmySuen/integral-human-pose\\
			TP-Net~\cite{rayat2018exploiting} & ECCV'18 & 58.2 & 44.1 & 22.0 & https://github.com/rayat137/Pose\_3D\\ 
			Liu et al.~\cite{liu2019feature} & TPAMI'19 & 61.1 & - & - & - \\
			EpipolarPose~\cite{Kocabas_2019_CVPR} & CVPR'19 & 51.8 & 45.0 & - & https://github.com/mkocabas/EpipolarPose \\
			VideoPose3D~\cite{3dvideopose} & CVPR'19 & 46.8 & 36.5 & 19.7&  https://github.com/facebookresearch/VideoPose3D\\
			SemGCN~\cite{zhao2019semantic} & CVPR'19 & 57.6 & - & - & https://github.com/garyzhao/SemGCN \\
			OANet~\cite{Cheng_2019_ICCV} & ICCV'19 & 42.9 & 32.8 & 14.3 & - \\
            Xu et al.~\cite{DKA_Xu_2020_CVPR} & CVPR'20 & 45.6 & 36.2 & 15.2 & - \\
			Liu et al.~\cite{liu2020attention} & CVPR'20 & 45.1 & 35.6 & 15.4 & https://github.com/lrxjason/Attention3DHumanPose \\
            \hline
            3DMPPE~\cite{moon2019camera} & ICCV'19 & 54.4 & - & - & https://github.com/mks0601/3DMPPE\_ROOTNET\_RELEASE \\
            Fabbri et al.~\cite{fabbri2020compressed} & CVPR'20 & 61.0 & 49.1 & - & https://github.com/fabbrimatteo/LoCO \\
			\hline
		\end{tabular}
	\end{table*}
	
	\begin{table*}
         \setlength\tabcolsep{4pt}
		\caption{Comparisons of 3D mesh recovery methods on Human3.6M, HumanEva, and 3DPW datasets.}
		\label{Comp_3dmesh}
		\centering
           	\footnotesize
		\begin{tabular}{c | c | c | c | c | c  | l }
			\hline
			\multirow{2}[1]{*}{\textbf{Method}} & \multirow{2}[1]{*}{\textbf{Pub.}}   & \multicolumn{2}{c|}{\textbf{Human3.6M}}&\multicolumn{1}{c|}{\textbf{HumanEva-I}}& \multicolumn{1}{c|}{\textbf{3DPW}}&\multirow{2}[1]{*}{\textbf{Code Link}}\\
			\cline{3-4}
			& & \textbf{MPJPE} & \textbf{PMPJPE} & \textbf{MPJPE }& \textbf{PMPJPE} \\
			\hline
			SMPLify~\cite{keep} & ECCV'16 & 82.3 & - & 79.9 & - & https://http://smplify.is.tue.mpg.de/\\
			UP~\cite{unite} & CVPR'17 & 80.7 & - & 74.5 & - & http:
			//up.is.tuebingen.mpg.de/\\
			HMR~\cite{hmr} & CVPR'18 & 87.9 & 58.1 & - & - &  https://github.com/akanazawa/hmr\\
			Human dynamics~\cite{kanazawa2018learning} & CVPR'19 & - & 56.9 & - & 72.6 & https://github.com/akanazawa/human\_dynamics\\
			GraphCMR~\cite{Kolotouros_2019_CVPR} & CVPR'19 & 71.9 & 50.1 & - & -& https://github.com/nkolot/GraphCMR \\
			HoloPose~\cite{Guler_2019_CVPR} & CVPR'19 & 60.2 & 46.5 & - & - & http://arielai.com/holopose\\
			DenseRaC~\cite{Xu_2019_ICCV} & ICCV'19 & 76.8 & 48.0 & - & -& - \\
			Texturepose~\cite{pavlakos2019texturepose} & ICCV'19 & - & 49.7 & - & -& https://github.com/geopavlakos/TexturePose \\
			Sun et al.~\cite{sun2019dsd-satn} & ICCV'19 & 59.1 & 42.4 & - & 69.5 & https://github.com/JDAI-CV/DSD\-SATN \\
			SPIN~\cite{kolotouros2019spin} & ICCV'19 & - & 41.1 & - & 59.2 & https://github.com/nkolot/SPIN\\
			VIBE~\cite{kocabas2019vibe} & CVPR'20 & 65.9 & 41.5 & - & 56.5 & https://github.com/mkocabas/VIBE\\
            CenterHMR~\cite{CenterHMR}  &arXiv'20& -& -& -& 53.2 & https://github.com/Arthur151/CenterHMR \\
			\hline
		\end{tabular}
         \vspace{-4mm}
	\end{table*}

    	\textbf{SURREAL Dataset~\cite{surreal}}
	is a large-scale synthetic dataset by rendering the textured SMPL model on the background images. The SMPL model is driven by numerous 3D motion capture data. However, the body textures are limited and low-resolution, which makes the rendered 2D images are un-realistic. 
    
  \textbf{AMASS~\cite{AMASS:2019}} is a large-scale motion capture (MoCap) dataset. It unifies 15 MoCap datasets by converting them to the SMPL parameters via MoSh++~\cite{AMASS:2019}. It contains more than 40 hours of motion data, spanning over 300 subjects, and more than 110K motions. AMASS is widely used to establish a prior human motion space by supervising the rationality of the estimated pose or motion.
  
 \textbf{$\bullet$}  \textbf{3D Multi-person Dataset}:
    
    \textbf{3DPW}~\cite{3dpw} is a single-view multi-person in-the-wild 3D human pose dataset that contains 60 video sequences (24 train, 24 test, and 12 validation) of rich activities, such as climbing, golfing, relaxing on the beach, etc. The videos are captured in various scenes, such as forest, street, playground, shopping mall, etc. They leverage IMU to obtain accurate 3D pose despite the complexity of scenes. Especially, 3DPW contains abundant 3D annotations, including 2D/3D pose annotations, 3D body scanning, and SMPL parameters. However, in some crowded scenes (e.g. on the street), 3DPW only provides the label of the target person, ignoring the pedestrians passing by. Generally, the entire dataset is used for evaluation, without any fine-tuning.
	
	\textbf{CMU Panoptic Dataset}~\cite{Joo_2017_TPAMI,Joo_2015_ICCV} is a large-scale multi-view and multi-person 3D pose dataset. Currently, it contains 65 sequences and 1.5 million 3D skeletons. They build an impressive dome for 360-degree motion capture, which contains 480 VGA cameras  (25 FPS), 31 HD cameras (30 FPS), 10 Kinect2 Sensors  (30 FPS), and 5 DLP Projectors. Especially, it contains multi-person social scenarios. Multi-person 3D pose estimation methods usually extract part of the data for evaluation. Zanfir et al.~\cite{zanfir2018monocular,zanfir2018deep} and Jiang et al.~\cite{Jiang_2020_CVPR} select 2 sub-sequences (9,600 frames from the HD camera
16 and 30) of 4 social activities (Haggling, Mafia, Ultimatum, and Pizza) for evaluation.

	\textbf{Joint Track Auto (JTA) Dataset~\cite{JTA}} 
	is a photo-realistic synthetic dataset for multi-person 3D pose evaluation. JTA is generated using the well-known video game \textit{Grand Theft Auto V}. It contains 512 HD videos of pedestrians walking in urban scenarios. Each video is 30s long and recorded at 30 FPS.

   For details of the widely used benchmarks, please refer to Table.~\ref{3D pose datasets}. Besides, we have released a detailed comparison and code toolbox for 3D pose data processing on Github\footnote{https://github.com/Arthur151/SOTA-on-monocular-3D-pose-and-shape-estimation}.

    \textbf{$\bullet$} \textbf{Analysis of Benchmark:}
    
    \textbf{Benchmark leaderboards} of 3D pose estimation and 3D mesh recovery are presented in Table.~\ref{Comp_3dpose} and Table.~\ref{Comp_3dmesh}, respectively. As we can see, the 3D pose estimation methods show obvious advantages in obtaining better 3D pose accuracy on indoor single-person 3D pose benchmarks, Human3.6M, and HumanEva. The 3D mesh recovery methods are more suitable for more comprehensive 3D human analysis and visualization. Besides, 3DPW~\cite{3dpw} is a new in-the-wild multi-person 3D pose benchmark. 3D mesh recovery methods have shown promising generalization ability on it.

    \textbf{Pose Space Analysis.}
    One of the main challenges for monocular 3D pose estimation is lacking sufficient training data, especially in diversity. Therefore, we analyze the pose space of four 3D benchmarks and visualize the statistical results. In detail, we first align the 3D pose annotations of all datasets to the pelvis, then perform cluster analysis. Here, we use the K-means clustering algorithm to evenly divide the pose space. For visualization, UMAP~\cite{mcinnes2018umap-software} is employed to reduce the dimension. The statistical results are drawn in Fig.~\ref{fig:pose_space}.

    The distribution of different activities in pose space and different clusters are shown in the first picture of Fig.~\ref{fig:pose_space}. 
    The sample density in pose space is very uneven. Most samples are gathered together. A similar conclusion could be drawn from the clustering results of the four benchmarks. We have observed that the 3D poses of most samples are close to walking or standing posture.
    The distribution of pose space is biased, which 
    limits the diversity of these datasets.

	\section{Conclusion and Future Directions.}
	This paper provides a comprehensive survey on deep learning-based monocular human pose estimation for both 2D and 3D tasks. We summarize over 200 papers on 2D and 3D tasks, according to the single-person and multi-person scenes based on image-level datasets or video-level datasets. Under some commonly used frameworks, human pose estimation methods have achieved significant progress via pose task-specific designs based on deep learning technologies. Despite great success, there are still challenges and many emerging topics that deserve further investigation and research. 
    
    
    \textbf{$\bullet$} \textbf{Pose Estimation for Complex Postures and Crowed Scenes.} 
    In daily life, the human body may make various complex or rare postures. 
    This makes the general models fail to accurately recognize the pose. In particular, for real-world applications, such as sports competitions for gymnastics, diving, and high jump, athletes may show extreme postures in a very short time. The complex and fast-changing postures will confuse the existing models that are trained on general datasets. Therefore, on the one hand, datasets for complex and rare postures are needed to improve the performance of the existing models. On the other hand, models with a stronger representation of complex behaviors and postures will be helpful. Additionally, in realistic scenarios, such as shopping malls, traffic surveillance, and sports events, pose estimation for crowded people is very challenging. Although some works~\cite{19MP_li2019crowdpose,cheng2020higherhrnet} have attempted to address crowds and occlusions in 2D pose estimation, they still suffer from low performance in real scenes. Moreover, for 3D pose estimation, occlusion caused by crowded scenes will confuse the models and cause unreasonable reconstructions for human shape and pose. Since the context of a complex scene contains clues for the interaction between person-person and person-object, further work may exploit the relation of scene and person to reason for invisible or occluded body parts. 
    
    \textbf{$\bullet$} \textbf{Benchmark, Protocol, and Toolkit for 3D Mesh Recovery.} Although monocular 3D human mesh recovery is a promising direction, due to the absence of large-scale 3D mesh datasets, the evaluation of 3D mesh recovery is usually performed on the skeleton-based 3D pose benchmarks, such as evaluating the MPJPE and PA-MPJPE for joint position error. Since the 3D mesh contains more information than the 3D keypoints, e.g., appearance information, this kind of indirect evaluation is not enough. Therefore, we need large-scale 3D human mesh datasets and protocols for comprehensive evaluations. Additionally, as the technology matures, industrial applications need easy-to-operate toolkits, especially for the lightweight implementation on cloud servers and hand-held devices. Some companies or communities, such as Google, Microsoft, and Tensorflow, have developed toolkits~\cite{google-toolkit,Microsoft-toolkit} or APIs (Application Programming Interface)~\cite{tensorflow-API} for 2D pose estimation. In the future, with the increasing application demands, we believe more mature and general toolboxes for both 2D and 3D pose estimation will further promote the implementation of the advanced algorithms.
    
	\textbf{$\bullet$} \textbf{Realistic Bodies with Expressive Faces, Hands, Hair, and Clothes.} 
	For better understanding the human in the scene, we need richer clues such as facial expression, emotional state, gesture, and clothes to describe a person in the 3D paradigm. Considering that the 3D pose recovery of a single body part (e.g., body, hand, and face) has been well developed in recent years, it is a natural tendency to move on to realistic 3D whole-body recovery with photorealistic details for hair, clothes, and expressive state. It is a brand new research field while only a few works~\cite{smplx,monocualar} make attempts to combine body, hands, and face in a unified representation framework. 
	The main challenges of this task lie in the lack of paired 3D pose datasets with all detailed information, and the scale differences between different body parts. Although there are many available separate 2D/3D body/face/hand/clothes/hair datasets, it is difficult to capture the realistic whole-body motion simultaneously. Therefore, further work may develop weak-/un-supervised methods to take advantage of all single-part data for effective learning. Moreover, with the development of computer graphics, photorealistic synthesized data may further improve the research. 
    
	 \textbf{$\bullet$} \textbf{Multi-person 3D Pose Estimation.} Although 2D pose estimation in multi-person scenes has been widely studied in recent years, the study of 3D cases has just begun. Multi-person 3D pose estimation is a very promising direction and is close to the real application scenarios. Most existing well-performing methods are multi-stage frameworks, which heavily rely on 2D human detection. Single-shot methods have the potential to achieve more attractive efficiency. However, when reasoning the poses especially the shapes of multiple people in the 3D form, the problems would be more complicated than in 2D keypoint estimation. For example, the interactive information of person-person or person-scene information is an important clue for determining the ambiguous poses, while it is usually ignored by most existing methods. Therefore, exploiting more comprehensive interactive information in the 2D images would be important for estimating more reasonable 3D poses.
	
	\textbf{$\bullet$} \textbf{Interaction with 3D World and Other Agents.}
	We live in a dynamic 3D world where people and objects interact with the environment. It would be interesting and promising to build an interaction-aware system that can capture and understand these embodied agents in the 3D world from monocular images. Although there have been some works focus on the 3D hand-object interaction~\cite{tekin2019ho,huang2020hot}, and the interaction of the 3D body with some specific objects~\cite{hassan2019resolving,zhang2020perceiving}, how to tackle the holistic 3D scene-understanding in uncontrolled in-the-wild scenes still remains challenging. On one hand, the 3D recovery of general objects from a monocular image has not been well solved. More detailed statistical models for general objects would bring a significant boost. On the other hand, instead of modeling the independent compositions, how to effectively represent the relationship of people, objects, and the scene will greatly affect the reasoning results. Moreover, with the development of 3D scene capture technologies, large-scale real-world 3D datasets would be transformative to boost the development of algorithms.  
	
	\textbf{$\bullet$} \textbf{Virtual Digital Human Generation with Emotion, Speech, and Communication.}
	A virtual digital person refers to a virtual person with a digital appearance character. It has specific features such as appearance, gender and personality, and the ability to express and communicate with language, facial expressions, and body movements. The technology has attracted lots of attention in many industries such as film production, virtual host, intelligent customer service, virtual teacher, etc.
	Modeling and generation of the 2D/3D human character is the core of virtual digital human products. Most of the existing products rely on computer graphics technologies and the marker motion capture equipment that is expensive and complex for operation. With the sharp increase in market demands, virtual digital humans are moving towards more intelligence, convenience, and diversified product forms.
	In the future, with the development of monocular 3D human recovery technology, the marker-less motion capture without professional sensing equipment is expected to realize simplicity, ease of use, and low price. Besides, although the current digital people have realized the intelligent synthesis of facial expression and mouth movement, the movements of other body parts only support recording and broadcasting. There will be more integrated and automated technology to realize the photorealistic 2D/3D whole-body model including the body movements, facial expressions, finger gestures, voices, etc. Additionally, the development of multimodal human-machine interaction will promote natural communication and interaction with digital people. 

\textbf{\textit{To summarize}}, monocular human pose estimation is a challenging and practical task. The development of deep learning for pose estimation is promising and exciting. In the future, both the research and application of human pose estimation contain many opportunities as well as challenges.  The future of monocular human pose estimation will largely depend on the practical focus and progress in algorithms, data, and application scenarios.

\ifCLASSOPTIONcaptionsoff
  \newpage
\fi



%
\bibliographystyle{IEEEtran}
\bibliography{Pose_Estimation_Survey}

%

\vfill
%

%




\end{document}